\documentclass[10pt,twocolumn,letterpaper]{article}

\usepackage[pagenumbers]{cvpr} 

\definecolor{cvprblue}{rgb}{0.21,0.49,0.74}
\usepackage[pagebackref,breaklinks,colorlinks,allcolors=cvprblue]{hyperref}
\usepackage[ruled]{algorithm2e}
\usepackage{multirow}
\usepackage{diagbox}
\usepackage{booktabs}
\usepackage{tabularx}

\def\paperID{18112} 
\def\confName{CVPR}
\def\confYear{2025}

\title{Self-Learning Hyperspectral and Multispectral Image Fusion via Adaptive Residual Guided Subspace Diffusion Model}

\author{
Jian Zhu, He Wang, Yang Xu\footnotemark[1], Zebin Wu\footnotemark[1], Zhihui Wei\\
School of Computer Science and Engineering, \\
Nanjing University of Science and Technology, Nanjing, China \\
{\tt\small \{123106010719,he\_wang,xuyangth90,wuzb,gswei\}@njust.edu.cn}
}

\begin{document}
\maketitle
\footnotetext[1]{Corresponding authors.}
\begin{abstract}

Hyperspectral and multispectral image (HSI-MSI) fusion involves combining a low-resolution hyperspectral image (LR-HSI) with a high-resolution multispectral image (HR-MSI) to generate a high-resolution hyperspectral image (HR-HSI). Most deep learning-based methods for HSI-MSI fusion rely on large amounts of hyperspectral data for supervised training, which is often scarce in practical applications. In this paper, we propose a self-learning Adaptive Residual Guided Subspace Diffusion Model (ARGS-Diff), which only utilizes the observed images without any extra training data. Specifically, as the LR-HSI contains spectral information and the HR-MSI contains spatial information, we design two lightweight spectral and spatial diffusion models to separately learn the spectral and spatial distributions from them. Then, we use these two models to reconstruct HR-HSI from two low-dimensional components, i.e, the spectral basis and the reduced coefficient, during the reverse diffusion process. Furthermore, we introduce an Adaptive Residual Guided Module (ARGM), which refines the two components through a residual guided function at each sampling step, thereby stabilizing the sampling process. Extensive experimental results demonstrate that ARGS-Diff outperforms existing state-of-the-art methods in terms of both performance and computational efficiency in the field of HSI-MSI fusion. Code is available at \href{https://github.com/Zhu1116/ARGS-Diff}{https://github.com/Zhu1116/ARGS-Diff}.

\end{abstract}    
\section{Introduction}
\label{sec:intro}

Hyperspectral images (HSI) are specialized images that contain extensive continuous spectral band information. Compared to natural images, HSIs provide more detailed spectral data, reflecting the intrinsic properties of various objects. This distinctive characteristic of HSIs makes them widely applicable across various domains, such as agricultural monitoring \cite{dale2013hyperspectral, khanal2020remote}, environmental protection \cite{li2020review}, urban planning \cite{huang2017multi}, and mineral exploration \cite{liu2013targeting, shirmard2022review}. However, due to hardware limitations, existing hyperspectral sensors often fail to deliver sufficient spatial resolution. To achieve higher spatial resolution, a common approach is HSI-MSI fusion, which involves fusing low-resolution hyperspectral images (LR-HSI) with high-resolution multispectral images (HR-MSI) to generate high-resolution hyperspectral images (HR-HSI) \cite{zhou2024general}. In the past decades, HSI-MSI fusion has garnered significant attention, and numerous related methods have emerged. 

Existing HSI-MSI fusion methods can be roughly categorized into two types, i.e, model-based methods \cite{shah2008efficient, loncan2015hyperspectral, yokoya2011coupled, simoes2014convex} and deep learning (DL)-based methods \cite{scarpa2018target, xie2020mhf, yao2020cross, wang2019deep}. Model-based approaches enhance image resolution by integrating spatial and spectral information from low-resolution hyperspectral images and high-resolution multispectral images. However, they rely on manually designed feature extractors and often assume independence between spatial and spectral information, which poses limitations in handling high-dimensional data and impedes the effective capture of complex features. In particular, when faced with the high dimensionality and nonlinearity of hyperspectral images, these methods struggle to deliver optimal performance and frequently require precise prior knowledge, thereby restricting their applicability.

In recent years, deep learning (DL) has made significant advancements in HSI-MSI fusion. These methods often rely on deep convolutional neural networks (DCNNs) \cite{wang2019deep, zhang2020ssr, yang2018hyperspectral} or Transformer networks \cite{vaswani2017attention, dosovitskiy2020image, wang2021pyramid, liu2021swin} to extract intricate spatial and spectral features from the images. DCNNs are particularly effective at capturing local patterns and hierarchical features, making them well-suited for spatial information extraction. On the other hand, Transformer networks excel at modeling long-range dependencies, allowing them to effectively capture global spectral information. While these methods outperform traditional model-based approaches in terms of feature extraction, they often rely heavily on large volumes of paired HSI-MSI data, which can be difficult to acquire in practical applications.
Recently, diffusion models (DMs) \cite{ho2020denoising, song2020score, miao2023dds2m,  dhariwal2021diffusion}, a novel generative models, have shown tremendous potential in image generation. With a more stable training process and superior generation quality, diffusion models have been widely applied in various image tasks, including image super-resolution \cite{saharia2022image, wang2022zero, xue2022ddrm}. In the realm of hyperspectral image fusion \cite{rui2023unsupervised, wu2023hsr, guo2023toward}, PLRDiff \cite{rui2023unsupervised} introduced a low-rank diffusion model tailored for hyperspectral image pan-sharpening, which leverages a pretrained deep diffusion model to provide image priors through Bayesian principles. In \cite{qu2024s2cyclediff}, a spatial-spectral-bilateral cycle diffusion framework (S2CycleDiff) was proposed  to progressively generate high-resolution hyperspectral images by learning a customized conditional cycle-diffusion network. However, despite these advancements, achieving high-quality results generally requires thousands of iterations, resulting in slow inference speeds and substantial computational costs. Some approaches \cite{song2020denoising, nichol2021improved, lu2022fast} have developed accelerated sampling strategies to reduce the number of iterations to a few dozen. Nevertheless, each inference step still demands substantial memory and floating-point computations, particularly in HSI-MSI fusion. Moreover, most edge devices (e.g., drones) are limited by storage and computational resources, making it challenging to deploy these demanding methods in practical remote sensing applications. Consequently, it is crucial to design a model that achieves low resource consumption and fast inference speeds while maintaining generation quality.

In this work, we propose a novel self-learning approach, the Adaptive Residual Guided Subspace Diffusion Model (ARGS-Diff). Based on the fact that HSI can be effectively reconstructed from the product of two low-dimensional components: the spectral basis and the reduced coefficient, we design and train two lightweight spectral and spatial networks and then use them to independently reconstruct the spectral basis and the reduced coefficient, respectively. The training process relies solely on the observed LR-HSI and HR-MSI to fully exploit the rich spectral and spatial information inherent in each of them. Furthermore, we introduce the Adaptive Residual Guided Module (ARGM) to ensure more stable convergence during the sampling process. In our method, both components need to be updated simultaneously at each step, and improper alignment between them may lead to instability or collapse. The ARGM alleviates this issue by applying a residual-guided function at each sampling step. It calculates the residual between the reconstructed two components and the target, then adjusts them to reduce the discrepancy, effectively guiding them toward better alignment with the true values. By incorporating this module, our approach demonstrates enhanced robustness to noise and achieves more stable and reliable reconstruction performance.

In summary, our contributions are as follows:
\begin{itemize}
    \item We propose a self-learning Adaptive Residual Guided Subspace Diffusion Model (ARGS-Diff), which effectively learns spectral and spatial distributions from the observed LR-HSI and HR-MSI using specially designed lightweight spectral and spatial networks, and reconstructs the HR-HSI from the estimated spectral basis and reduced coefficient.
    \item We introduce the Adaptive Residual Guided Module (ARGM), which refines the spectral and spatial components through a residual-guided function, effectively enhancing the stability and convergence of the sampling process.
    \item Extensive experiments conducted on four datasets demonstrate the superior performance of the proposed method over existing state-of-the-art approaches.
\end{itemize}
\section{Preliminaries}
\label{sec:pre}











\subsection{Diffusion Models}

The diffusion model reconstructs data by progressively adding noise during a forward process and learning the reverse denoising pathway to recover the original data \cite{ho2020denoising, song2020denoising}. Typically, a diffusion process is composed of a forward process with $T$ steps of noise addition, followed by a reverse process over the same $T$ steps. Given a noise schedule $\{\bar{\alpha}_t\}_{t=1}^T$ and clean input data $\mathbf{x}_0$, the noise addition in each step is defined by:

\begin{equation}
\label{eq1}
\mathbf{x}_t = \sqrt{\bar{\alpha}_t} \mathbf{x}_0 + (\sqrt{1 - \bar{\alpha}_t}) \epsilon, \quad \epsilon \sim \mathcal{N}(0, I).
\end{equation}

\noindent The model is trained to predict the noise $\epsilon$ by minimizing the error between the estimated and true noise:

\begin{equation}
\epsilon_{\theta}(\mathbf{x}_t, t) \approx \epsilon = \frac{\mathbf{x}_t - \sqrt{\bar{\alpha}_t} \mathbf{x}_0}{\sqrt{1 - \bar{\alpha}_t}}.
\end{equation}

\noindent In the reverse process, starting from random noise, the model iteratively refines the estimate over \( T \) steps. At each step \( t \), the model estimates the clean input $\hat{\mathbf{x}}_0$ as:

\begin{equation}
\label{eq3}
\hat{\mathbf{x}}_0 = \frac{\mathbf{x}_t - (1 - \bar{\alpha}_t) \epsilon_{\theta}(\mathbf{x}_t, t)}{\sqrt{\bar{\alpha}_t}}.
\end{equation}

\noindent Subsequently, $\mathbf{x}_{t-1}$ is computed by:

\begin{equation}
\mathbf{x}_{t-1} = \sqrt{\bar{\alpha}_{t-1}} \hat{\mathbf{x}}_0 + (\sqrt{1 - \bar{\alpha}_{t-1}}) \epsilon_{\theta}.
\end{equation}

\noindent The final reconstructed output $\mathbf{x}_0$ is obtained after all reverse steps are completed.


\begin{figure*}[htbp]
    \centering
    \includegraphics[width=0.99\textwidth]{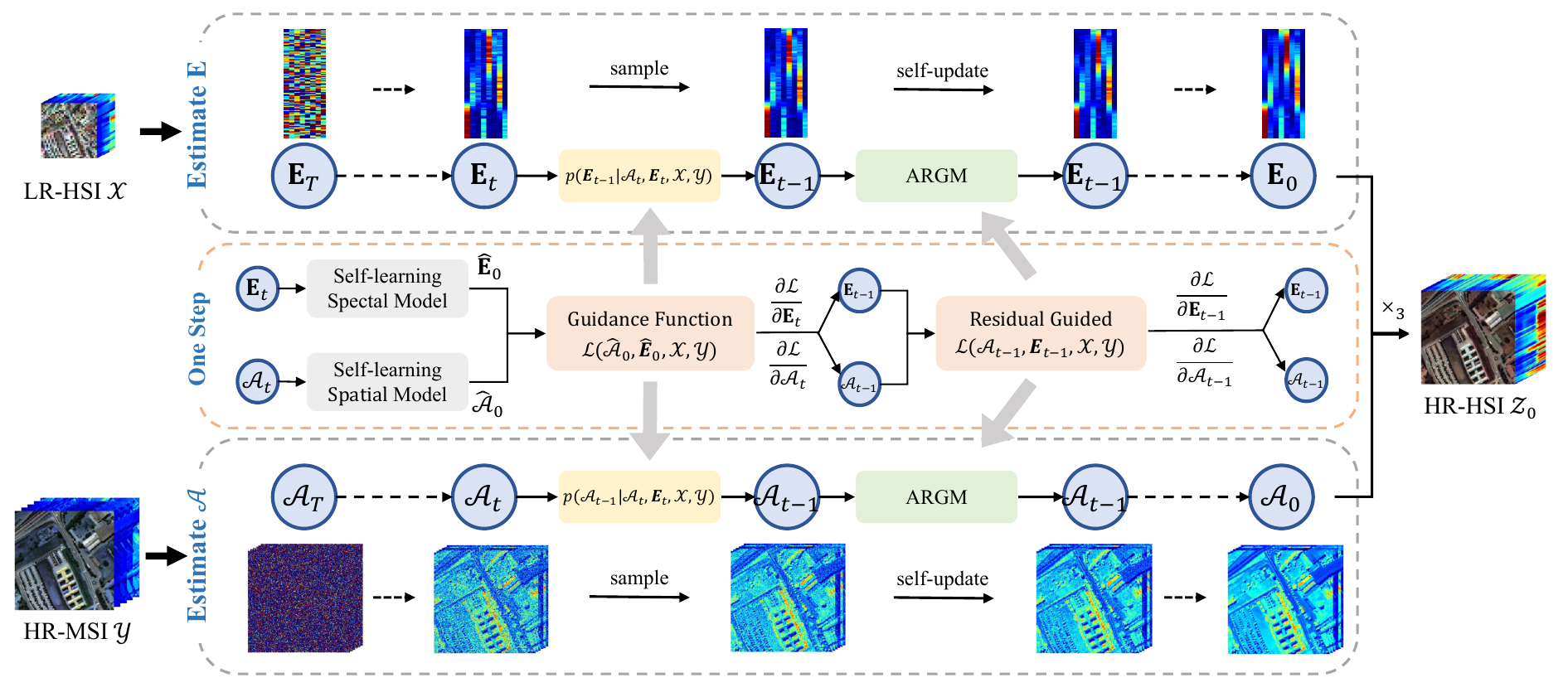}
    \caption{The overall framework of the proposed ARGS-Diff.}
    \label{fig1}
\end{figure*}

\subsection{HSI-MSI Fusion}

In HSI-MSI fusion problem, the input data consists of a LR-HSI $\mathcal{X} \in \mathbb{R}^{h \times w \times C}$ and a HR-MSI $\mathcal{Y} \in \mathbb{R}^{H \times W \times c}$, which are fused to produce a HR-HSI $\mathcal{Z} \in \mathbb{R}^{H \times W \times C}$. Typically, the relationship between the LR-HSI, HR-MSI, and HR-HSI is represented through degradation models as follows:

\begin{equation}
\mathcal{X} = \mathcal{H}(\mathcal{Z}) + \mathcal{N}_\mathcal{X},
\end{equation}
\begin{equation}
\mathcal{Y} = \mathcal{Z} \times_3 \mathbf{R} + \mathcal{N}_\mathcal{Y},
\end{equation}

\noindent where $\mathcal{H}(\cdot)$ represents downsampling in the spatial domain, $\mathbf{R}$ is the spectral response function (SRF) matrix, $\times_3$ denotes mode-3  multiplication, $\mathcal{N}_\mathcal{X}$ and $\mathcal{N}_\mathcal{Y}$ are the measurements noise, which is assumed to be the Gaussian noise.

A common approach for HSI-MSI fusion is to decompose the HSI into two low-dimensional components, process each separately, and then combine them to synthesize a clean HR-HSI. The decomposition is expressed as:

\begin{equation}
\label{decom}
\mathcal{Z} = \mathcal{A} \times_3 \mathbf{E},
\end{equation}

\noindent where $\mathcal{A} \in \mathbb{R}^{H \times W \times d}$ represents the reduced coefficient, $\mathbf{E} \in \mathbb{R}^{d \times C}$ represents the spectral basis, $d$ is the number of subspace dimensions, and $d \ll C$.

Compared to the high dimensional of HSI, these two low-dimensional components have reduced data complexity, making them easier to process. In this paper, we follow this approach by training a spectral and spatial network using the LR-HSI and HR-MSI. During the sampling process, these networks separately reconstruct the corresponding spectral basis and reduced coefficient. The final fused HR-HSI is obtained from the product of these two components.

\section{Proposed Method}
\label{sec:propose}

\subsection{Overview}

As described in Eq. (\ref{decom}), the HSI can be decomposed into two low-dimensional components, i.e, the spectral basis and the reduced coefficient. We train two separate networks to reconstruct these components, which are then combined to form the final HR-HSI. 
Specifically, We design a spectral network and a spatial network, and train them using the LR-HSI and HR-MSI based on the forward process of the diffusion model to separately extract the spectral and spatial distribution information. The detailed network architecture and training procedure are described in Section \ref{sec:3.2}. 

During the sampling process, we leverage the posterior sampling theory \cite{chung2022diffusion} to iteratively reconstruct the spectral basis and reduced coefficient using the well-trained networks, conditioning on the observed LR-HSI and HR-MSI. The detailed sampling process is outlined in Section \ref{sec:3.3}. Additionally, since we need to update two components simultaneously at each sampling step, which may lead to instability if they are not properly aligned, we introduce the Adaptive Residual Guided Module (ARGM) to iteratively refine the spectral and spatial components at each step, ensuring the stability of the sampling process, as described in Section \ref{sec:3.4}. After $T$ sampling steps, the HR-HSI is obtained by taking the product of the reconstructed spectral basis and reduced coefficient.

\subsection{Self-learning Subspace Networks}
\label{sec:3.2}

Recent advancements in diffusion models have introduced many techniques, such as attention mechanisms, to enhance performance. However, these innovations often come with a significant increase in network size, resulting in higher resource consumption and longer image generation times. To address this challenge, we design two lightweight networks specifically tailored for hyperspectral image fusion: a spectral network and a spatial network. This design significantly reduces sampling time without compromising performance in our method.

The spectral network is a fully connected network consisting of five layers, where the input and output feature dimensions match the number of spectral bands, \(C\), in the LR-HSI. The spatial network adopts a "UNet-like" architecture \cite{ronneberger2015u}, comprising nine convolutional layers: four downsampling layers, one intermediate layer, and four upsampling layers. Each convolutional layer includes two residual blocks, and the input and output channels are set to the number of subspace dimensions, \(d\).

To optimize network training, we restructure the training samples. For the spectral network, LR-HSI, which contains abundant spectral information, is leveraged to represent the spectral distribution of the HR-HSI. Specifically, we randomly select \( d \) pixel spectra from the LR-HSI and concatenate them into a sample of size \( (d, C) \) for the spectral network's input. For the spatial network, HR-MSI, which retains rich spatial information, is used for training. We extract a patch from the HR-MSI, randomly select one band from \(c\) spectral bands within the patch, and repeat this process \( d \) times to form a sample of size \( (H_{\text{patch}}, W_{\text{patch}}, d) \). This approach not only facilitates effective training but also introduces diversity into the input samples, enabling the network to better capture the spatial and spectral features of the images.

\subsection{Subspace Reverse Diffusion Process}
\label{sec:3.3}

Recently, several methods \cite{bansal2023universal, chung2022diffusion} have been proposed to generate images using various conditional inputs. In general, given a condition \( \mathbf{y} \), these methods model the reverse sampling process as \( p_{\theta}(\mathbf{x}_{t-1} | \mathbf{x}_t, \mathbf{y}) \) incorporating a gradient term into the noise estimation function \( \epsilon_{\theta} \). This is formulated as:
\begin{equation}
\label{eq8}
\hat{\epsilon}_\theta(\mathbf{x}_t, t) = \epsilon_\theta(\mathbf{x}_t, t) - \rho \nabla_{\mathbf{x}_t} \mathcal{L}(\hat{\mathbf{x}}_0, y),
\end{equation}
where \( \rho \) denotes the step size, and \( \mathcal{L} \) is a guidance function that measures the distance between the predicted image \( \hat{\mathbf{x}}_0 \) and the target image. In our case, we use the LR-HSI \( \mathcal{X} \) and HR-MSI \( \mathcal{Y} \) as conditions, employing spatial and spectral networks to estimate the spectral basis and the reduced coefficient. Specifically, Eq. (\ref{eq8}) can be rewritten as:
\begin{equation}
\begin{aligned}
\label{f:9}
&\hat{s}_\theta(\mathcal{A}_t, t) = s_\theta(\mathcal{A}_t, t) - \rho_1 \nabla_{\mathcal{A}_t} \mathcal{L}(\hat{\mathcal{A}}_0, \hat{\mathbf{E}}_0, \mathcal{X}, \mathcal{Y}), \\
&\hat{c}_\zeta(\mathbf{E}_t, t) = c_\zeta(\mathbf{E}_t, t) - \rho_2 \nabla_{\mathbf{E}_t} \mathcal{L}(\hat{\mathcal{A}}_0, \hat{\mathbf{E}}_0, \mathcal{X}, \mathcal{Y}),
\end{aligned}
\end{equation}
where \( s_\theta(\cdot) \) and \( c_\zeta(\cdot) \) represent the spatial and spectral networks, respectively, $\rho_1$ and $\rho_2$ are the step size for updating $s_\theta(\mathcal{A}_t, t)$ and $c_\zeta(\mathbf{E}_t, t)$. The guidance function is defined as:
\begin{equation}
\label{L_1}
\begin{aligned}
\mathcal{L}(\hat{\mathcal{A}}_0, \hat{\mathbf{E}}_0, \mathcal{X}, \mathcal{Y}) & =\ \|\mathcal{H}(\hat{\mathcal{A}}_{0} \times_3 \hat{\mathbf{E}}_{0}) - \mathcal{X}\|^2_2 \\ & + \lambda_1 \|\hat{\mathcal{A}}_{0} \times_3 \hat{\mathbf{E}}_{0} \times_3 \mathbf{R} - \mathcal{Y}\|^2_2,
\end{aligned}
\end{equation}
where $\lambda_1$ is weight to balance the two observations. \( \hat{\mathcal{A}}_0 \) and \( \hat{\mathbf{E}}_0 \) can be estimated from Eq. (\ref{eq3}) as:
\begin{equation}
\label{hat_0}
\begin{aligned}
&\hat{\mathcal{A}}_0 = \frac{\mathcal{A}_t - (\sqrt{1 - \bar{\alpha}_t}) s_\theta(\mathcal{A}_t, t)}{\sqrt{\bar{\alpha}_t}}, \\
&\hat{\mathbf{E}}_0 = \frac{\mathbf{E}_t - (\sqrt{1 - \bar{\alpha}_t}) c_\zeta(\mathbf{E}_t, t)}{\sqrt{\bar{\alpha}_t}}.
\end{aligned}
\end{equation}
Next, \( \mathcal{A}_{t-1} \) and \( \mathbf{E}_{t-1} \) are sampled from Eq. (\ref{eq1}) as:
\begin{equation}
\label{t-1}
\begin{aligned}
&\mathcal{A}_{t-1} = \sqrt{\bar{\alpha}_{t-1}} \hat{\mathcal{A}}_0 + (\sqrt{1 - \bar{\alpha}_{t-1}}) \hat{s}_\theta(\mathcal{A}_t, t), \\
&\mathbf{E}_{t-1} = \sqrt{\bar{\alpha}_{t-1}} \hat{\mathbf{E}}_0 + (\sqrt{1 - \bar{\alpha}_{t-1}}) \hat{c}_\zeta(\mathbf{E}_t, t).
\end{aligned}
\end{equation}

\noindent After \( \mathcal{A}_{t-1} \) and \( \mathbf{E}_{t-1} \) are sampled from Eq. (\ref{t-1}) at current step, they are further refined using the proposed ARGM, as detailed in Section \ref{sec:3.4}. Upon completion of the sampling process, the reconstructed HR-HSI \( \mathcal{Z}_0 \) is derived from \( \mathcal{A}_0 \times_3 \mathbf{E}_0 \). The proposed method is summarized in Algorithm \ref{Alg}.

\textbf{Remark:} To accelerate convergence, we integrate the Adam optimizer \cite{kingma2014adam} into Eq. (\ref{f:9}) with the following updates:
\begin{equation}
\label{mv}
\begin{aligned}
&\mathbf{m}_{t-1}^{(\mathcal{A})} = \beta_1 \mathbf{m}_t^{(\mathcal{A})} + (1 - \beta_1) \nabla_{\mathcal{A}_t} \mathcal{L}(\hat{\mathcal{A}}_0, \hat{\mathbf{E}}_0, \mathcal{X}, \mathcal{Y}), \\
&\mathbf{v}_{t-1}^{(\mathcal{A})} = \beta_2 \mathbf{v}_t^{(\mathcal{A})} + (1 - \beta_2) \nabla_{\mathcal{A}_t} \mathcal{L}(\hat{\mathcal{A}}_0, \hat{\mathbf{E}}_0, \mathcal{X}, \mathcal{Y})^2,\\
&\mathbf{m}_{t-1}^{(\mathbf{E})} = \beta_1 \mathbf{m}_t^{(\mathbf{E})} + (1 - \beta_1) \nabla_{\mathbf{E}_t} \mathcal{L}(\hat{\mathcal{A}}_0, \hat{\mathbf{E}}_0, \mathcal{X}, \mathcal{Y}), \\
&\mathbf{v}_{t-1}^{(\mathbf{E})} = \beta_2 \mathbf{v}_t^{(\mathbf{E})} + (1 - \beta_2) \nabla_{\mathbf{E}_t} \mathcal{L}(\hat{\mathcal{A}}_0, \hat{\mathbf{E}}_0, \mathcal{X}, \mathcal{Y})^2,
\end{aligned}
\end{equation}
where \( \mathbf{m}_{t-1} \) and \( \mathbf{v}_{t-1} \) represent the first and second moment estimates in the Adam optimizer, with superscripts \((\mathcal{A})\) and \((\mathbf{E})\) denoting the corresponding spatial and spectral matrices. The parameters \( \beta_1 \) and \( \beta_2 \) are the decay rates for these moment estimates.

We then rectify the first and second moment estimates as:
\begin{equation}
\label{mvhat}
\begin{aligned}
&\hat{\mathbf{m}}_{t-1}^{(\mathcal{A})} = \frac{\mathbf{m}_{t-1}^{(\mathcal{A})}}{1 - \beta_1^{T-t}},
\hat{\mathbf{v}}_{t-1}^{(\mathcal{A})} = \frac{\mathbf{v}_{t-1}^{(\mathcal{A})}}{1 - \beta_2^{T-t}},\\
&\hat{\mathbf{m}}_{t-1}^{(\mathbf{E})} = \frac{\mathbf{m}_{t-1}^{(\mathbf{E})}}{1 - \beta_1^{T-t}},
\hat{\mathbf{v}}_{t-1}^{(\mathbf{E})} = \frac{\mathbf{v}_{t-1}^{(\mathbf{E})}}{1 - \beta_2^{T-t}}.
\end{aligned}
\end{equation}

\noindent Finally, we rewrite Eq. (\ref{f:9}) as:
\begin{equation}
\label{sc}
\begin{aligned}
&\hat{s}_\theta(\mathcal{A}_t, t) = s_\theta(\mathcal{A}_t, t) - \rho_1 \frac{\hat{\mathbf{m}}_{t-1}^{(\mathcal{A})}}{\sqrt{\hat{\mathbf{v}}_{t-1}^{(\mathcal{A})}} + \epsilon}, \\
&\hat{c}_\zeta(\mathbf{E}_t, t) = c_\zeta(\mathbf{E}_t, t) - \rho_2 \frac{\hat{\mathbf{m}}_{t-1}^{(\mathbf{E})}}{\sqrt{\hat{\mathbf{v}}_{t-1}^{(\mathbf{E})}} + \epsilon},
\end{aligned}
\end{equation}

\noindent where \( \epsilon \) is a small constant introduced to avoid division by zero.

\subsection{ARGM}
\label{sec:3.4}

Unlike existing diffusion posterior sampling methods \cite{chung2022diffusion, rui2023unsupervised}, which update only a single intermediate sample at each sampling step, our approach requires simultaneous updates of both the spectral basis and the reduced coefficient. This introduces additional instability into the sampling process, as both components must be adjusted to align with each other in each iteration. To alleviate this issue, we introduce the Adaptive Residual Guided Module (ARGM), which facilitates better alignment between the spectral and spatial components, resulting in more stable convergence.

Specifically, after obtaining \( \mathcal{A}_{t-1} \) and \( \mathbf{E}_{t-1} \) at the \( t \)-th sampling step, we compute a residual loss that quantifies the discrepancy between the reconstructed two components and the target to refine them. The residual loss can be calculated by:

\begin{equation}
\label{L_2}
\begin{aligned}
\mathcal{L}(\mathcal{A}_{t-1}, \mathbf{E}_{t-1}, \mathcal{X}, \mathcal{Y}) & = \ 
\|\mathcal{H}(\mathcal{A}_{t-1} \times_3 \mathbf{E}_{t-1}) - \mathcal{X}\|^2_2 \\
& + \lambda_2 \|\mathcal{A}_{t-1} \times_3 \mathbf{E}_{t-1} \times_3 \mathbf{R} - \mathcal{Y}\|^2_2,
\end{aligned}
\end{equation}

\noindent where \( \lambda_2 \) is a weight balancing the two objectives. The first term ensures that the predicted LR-HSI \( \mathcal{H}(\mathcal{A}_{t-1} \times_3 \mathbf{E}_{t-1}) \) is consistent with the observed image \( \mathcal{X} \), while the second term penalizes the discrepancy between the predicted HR-MSI ($\mathcal{A}_{t-1} \times_3 \mathbf{E}_{t-1} \times_3 \mathbf{R}$) and the target \( \mathcal{Y} \). This ensures that the spectral and spatial components are aligned both individually and in combination.

Then, the spatial and spectral components are updated by:

\begin{equation}
\label{t-1_}
\begin{aligned}
&\mathcal{A}_{t-1} := \mathcal{A}_{t-1} - \frac{\rho_1}{r} \nabla_{\mathcal{A}_{t-1}} \mathcal{L}(\mathcal{A}_{t-1}, \mathbf{E}_{t-1}, \mathcal{X}, \mathcal{Y}), \\
&\mathbf{E}_{t-1} := \mathbf{E}_{t-1} - \frac{\rho_2}{r} \nabla_{\mathbf{E}_{t-1}} \mathcal{L}(\mathcal{A}_{t-1}, \mathbf{E}_{t-1}, \mathcal{X}, \mathcal{Y}),
\end{aligned}
\end{equation}

\noindent where \( r \) is a ratio that controls the relative step sizes for updating the spatial component and the spectral component, respectively.

In this way, the ARGM computes the residual between the reconstructed two components and the target at each step, and then uses the residual-guided function to adjust both components. On the one hand, this brings the components closer to the target direction, reducing the instability that arises from updating both components simultaneously. On the other hand, it compensates for the prediction errors introduced by the model at each sampling step, thus making the model more robust and less sensitive to noise and discrepancies.

By iteratively refining the spatial and spectral components using residual-guided function, the ARGM helps stabilize the convergence of the diffusion sampling process, leading to improved performance and robustness in the HSI-MSI fusion task.

\begin{algorithm}
    \caption{Reverse Diffusion Process of ARGS-Diff}
    \label{Alg}
    \KwIn{LR-HSI \(\mathcal{X}\), HR-MSI \(\mathcal{Y}\), \(\mathcal{A}_T\) and \(\mathbf{E}_T\) sampled from \(\mathcal{N}(0,I)\), spatial network \(s_\theta\), spectral network \(c_\zeta\), balance weight \(\lambda_1\), \(\lambda_2\), step size \(\rho_1\), \(\rho_2\), ratio \(r\), Adam params \(\mathbf{m}_T^{(\mathcal{A})}\), \(\mathbf{v}_T^{(\mathcal{A})}\), \(\mathbf{m}_T^{(\mathbf{E})}\), \(\mathbf{v}_T^{(\mathbf{E})}\), \(\beta_1\), \(\beta_2\), \(\epsilon\)}
    \KwOut{HR-HSI \(\mathcal{Z}_0\)}

    Initialize \(\mathbf{m}_T^{(\mathcal{A})}=\mathbf{v}_T^{(\mathcal{A})}=\mathbf{m}_T^{(\mathbf{E})}=\mathbf{v}_T^{(\mathbf{E})}=\mathbf{0}\)
    
    \For{\(t = T, T-1,\cdots 1\)}
    {
        \textbf{step 1:} estimate \(\hat{\mathcal{A}}_0\), \(\hat{\mathbf{E}}_0\) by Eq. (\ref{hat_0}) \\
        \textbf{step 2:} calculate \(\mathcal{L}(\hat{\mathcal{A}}_0, \hat{\mathbf{E}}_0, \mathcal{X}, \mathcal{Y})\) by Eq.(\ref{L_1}) \\
        \textbf{step 3:} calculate \(\hat{\mathbf{m}}_{t-1}^{(\mathcal{A})}\), \(\hat{\mathbf{v}}_{t-1}^{(\mathcal{A})}\), \(\hat{\mathbf{m}}_{t-1}^{(\mathbf{E})}\), \(\hat{\mathbf{v}}_{t-1}^{(\mathbf{E})}\) by Eq. (\ref{mv}), Eq. (\ref{mvhat}) \\ 
        \textbf{step 4:} estimate  \(\hat{s}_\theta(\mathcal{A}_t, t)\), \(\hat{c}_\zeta(\mathbf{E}_t, t)\) by Eq. (\ref{sc}) \\
        \textbf{step 5:} sample \(\mathcal{A}_{t-1}\), \(\mathbf{E}_{t-1}\) by Eq. (\ref{t-1}) \\
        \textbf{step 6:} calculate \(\mathcal{L}(\mathcal{A}_{t-1}, \mathbf{E}_{t-1}, \mathcal{X}, \mathcal{Y})\) by Eq. (\ref{L_2}) \\
        \textbf{step 7:} update \(\mathcal{A}_{t-1}\), \(\mathbf{E}_{t-1}\) by Eq. (\ref{t-1_}) \\
    }
    
    \textbf{return} \(\mathcal{Z}_0 = \mathcal{A}_0 \times_3 \mathbf{E}_0\)
\end{algorithm}
\section{Experiment}
\label{sec:experiment}

\subsection{Datasets and Evaluation Metrics}

We conducted a series of simulated experiments using three public datasets to assess the performance of the proposed method for HSI fusion: Pavia University, Chikusei, and KSC. The original dimensions of these datasets are \(610 \times 340 \times 103\), \(2517 \times 2335 \times 128\), and \(512 \times 614 \times 176\). We cropped the central regions of each to obtain HSIs with sizes \(256 \times 256 \times 103\), \(256 \times 256 \times 128\), and \(256 \times 256 \times 176\). The LR-HSI was generated by downsampling the HR-HSI using bicubic interpolation with a scale factor of 4, and the HR-MSI was obtained by performing mode-3 tensor multiplication of the HR-HSI with a simulated spectral response function. Additionally, to evaluate the method’s real-world applicability, we conducted experiments on the DFC2018 Houston dataset, which provides real hyperspectral data. The following four commonly used evaluation metrics were employed to assess model performance: peak signal-to-noise ratio (PSNR), spectral angle mapper (SAM), relative dimensionless global error in synthesis (ERGAS) , and structural similarity (SSIM).

\subsection{Implementation Details}

All experiments were carried out on an NVIDIA GeForce RTX 4090 GPU. To test the robustness of the model to noise, we added noise to all LR-HSI and HR-MSI inputs with a Signal-to-Noise Ratio (SNR) of 35. The diffusion sampling step \(T\) was set to 500, using an exponential noise schedule \cite{pang2024hir}. The balance weights \(\lambda_1\) and \(\lambda_2\) were both set to 1, while the step sizes \(\rho_1\) and \(\rho_2\) were set to 0.05, and the ratio \(r\) was set to 10. The Adam optimizer was used with \(\beta_1 = 0.9\) and \(\beta_2 = 0.999\), and \(\epsilon = 1 \times 10^{-8}\). The number of subspace dimensions \(d\) was set to 8.

\subsection{Experimental Results}
\subsubsection{Results on Simulated Data}

We compare the proposed method with six well-established approaches, including traditional methods such as CNMF \cite{yokoya2011coupled} and HySure \cite{simoes2014convex}, deep learning-based methods like CuCaNet \cite{yao2020cross} and MIAE \cite{liu2022model}, and diffusion model-based approaches such as PLRDiff \cite{rui2023unsupervised} and S$^2$CycleDiff \cite{qu2024s2cyclediff}. The quantitative results are summarized in Tables \ref{tab:1}, \ref{tab:2}, and \ref{tab:3}. From these results, it is clear that the proposed method consistently outperforms all other competing approaches across all evaluation metrics on three different datasets. In terms of PSNR, the proposed method achieves the highest values, surpassing the second-best method, MIAE, by 1.31 dB, 1.37 dB, and 1.27 dB on the respective datasets. This demonstrates the superior ability of ARGS-Diff to reconstruct high-fidelity hyperspectral images, preserving both spectral and spatial details with greater accuracy. When examining the computational efficiency, traditional methods such as CNMF and HySure offer faster inference speeds, but they result in poorer fusion performance. Deep learning methods like CuCaNet and MIAE show improved accuracy, but at the cost of longer processing times. Diffusion model-based methods like PLRDiff and S$^2$CycleDiff also demand significant inference time, while the proposed method achieves high-quality fusion results with a reduced runtime. 

\begin{figure*}
  \centering

  \begin{subfigure}{0.107\textwidth}
    \includegraphics[width=\linewidth]{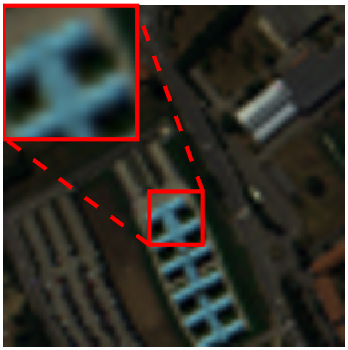}
  \end{subfigure}
  \hfill
  \begin{subfigure}{0.107\textwidth}
    \includegraphics[width=\linewidth]{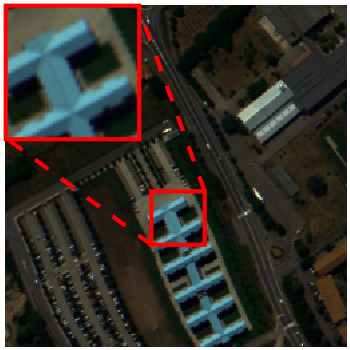}
  \end{subfigure}
  \hfill
  \begin{subfigure}{0.107\textwidth}
    \includegraphics[width=\linewidth]{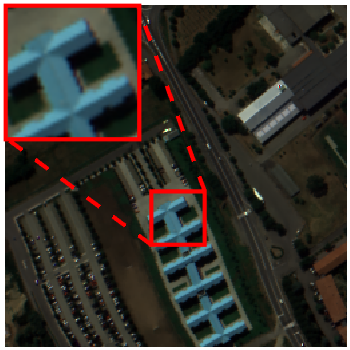}
  \end{subfigure}
  \hfill
  \begin{subfigure}{0.107\textwidth}
    \includegraphics[width=\linewidth]{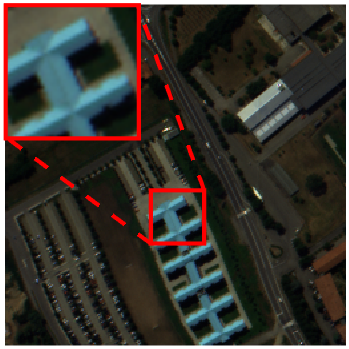}
  \end{subfigure}
  \hfill
  \begin{subfigure}{0.107\textwidth}
    \includegraphics[width=\linewidth]{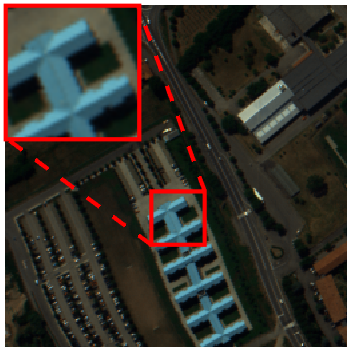}
  \end{subfigure}
  \hfill
  \begin{subfigure}{0.107\textwidth}
    \includegraphics[width=\linewidth]{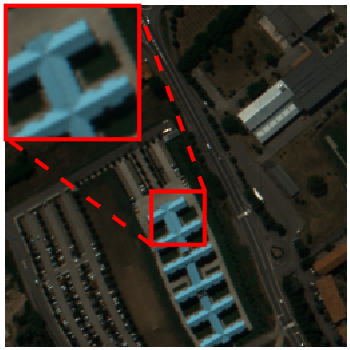}
  \end{subfigure}
  \hfill
  \begin{subfigure}{0.107\textwidth}
    \includegraphics[width=\linewidth]{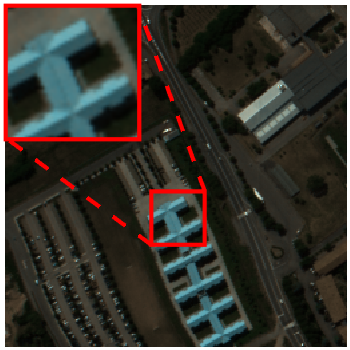}
  \end{subfigure}
  \hfill
  \begin{subfigure}{0.107\textwidth}
    \includegraphics[width=\linewidth]{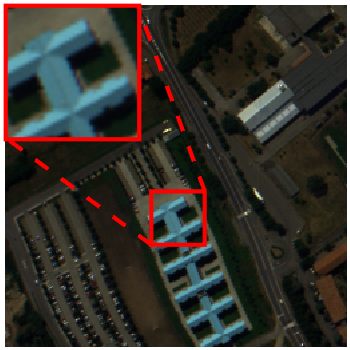}
  \end{subfigure}
  \hfill
  \begin{subfigure}{0.107\textwidth}
    \includegraphics[width=\linewidth]{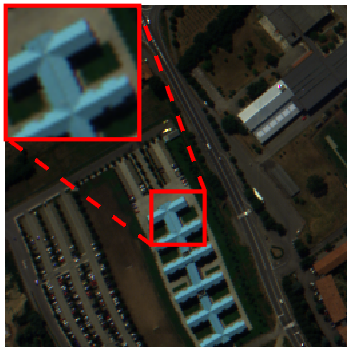}
  \end{subfigure}

  \vspace{0.2mm}

  \begin{subfigure}{0.107\textwidth}
    \includegraphics[width=\linewidth]{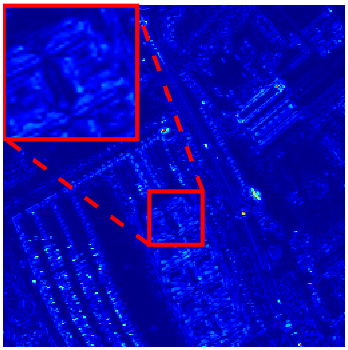}
  \end{subfigure}
  \hfill
  \begin{subfigure}{0.107\textwidth}
    \includegraphics[width=\linewidth]{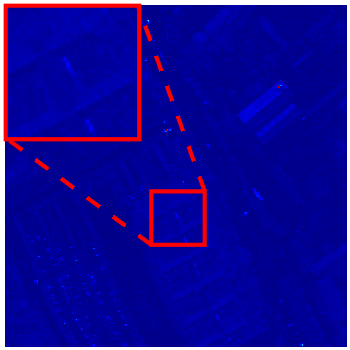}
  \end{subfigure}
  \hfill
  \begin{subfigure}{0.107\textwidth}
    \includegraphics[width=\linewidth]{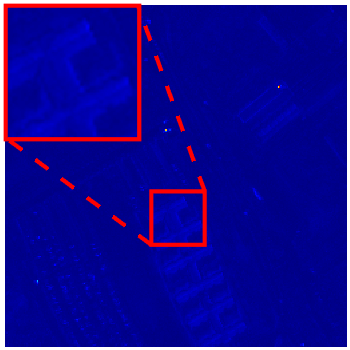}
  \end{subfigure}
  \hfill
  \begin{subfigure}{0.107\textwidth}
    \includegraphics[width=\linewidth]{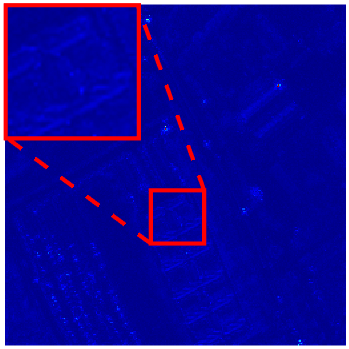}
  \end{subfigure}
  \hfill
  \begin{subfigure}{0.107\textwidth}
    \includegraphics[width=\linewidth]{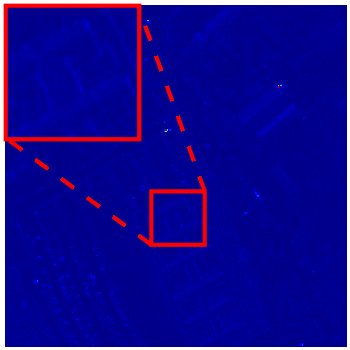}
  \end{subfigure}
  \hfill
  \begin{subfigure}{0.107\textwidth}
    \includegraphics[width=\linewidth]{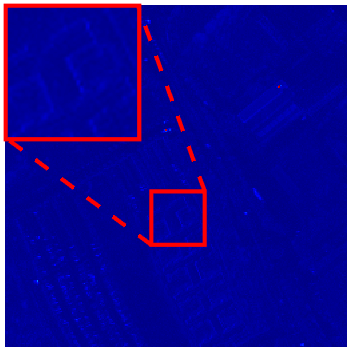}
  \end{subfigure}
  \hfill
  \begin{subfigure}{0.107\textwidth}
    \includegraphics[width=\linewidth]{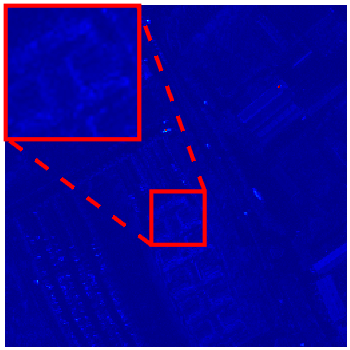}
  \end{subfigure}
  \hfill
  \begin{subfigure}{0.107\textwidth}
    \includegraphics[width=\linewidth]{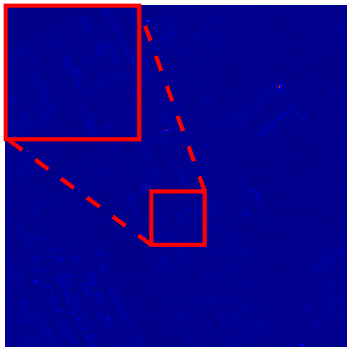}
  \end{subfigure}
  \hfill
  \begin{subfigure}{0.107\textwidth}
    \includegraphics[width=\linewidth]{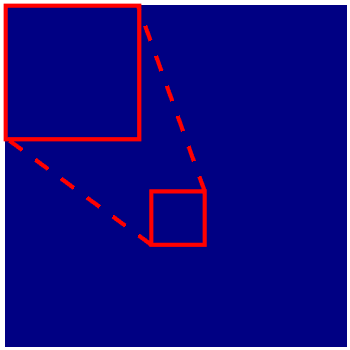}
  \end{subfigure}

  \vspace{0.2mm}

  \begin{subfigure}{0.107\textwidth}
    \includegraphics[width=\linewidth]{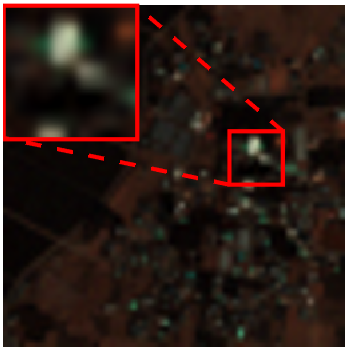}
  \end{subfigure}
  \hfill
  \begin{subfigure}{0.107\textwidth}
    \includegraphics[width=\linewidth]{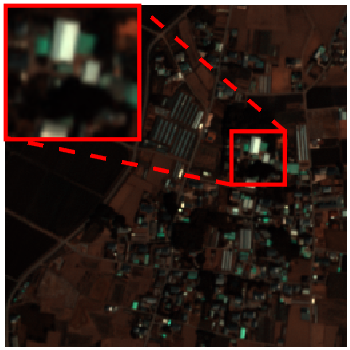}
  \end{subfigure}
  \hfill
  \begin{subfigure}{0.107\textwidth}
    \includegraphics[width=\linewidth]{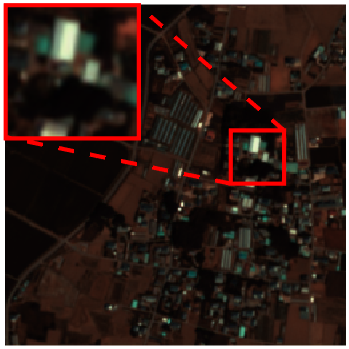}
  \end{subfigure}
  \hfill
  \begin{subfigure}{0.107\textwidth}
    \includegraphics[width=\linewidth]{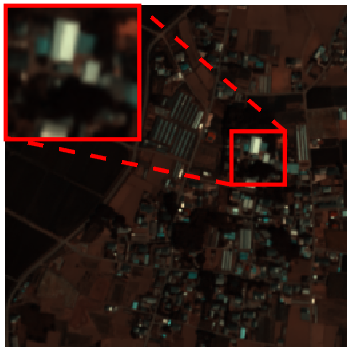}
  \end{subfigure}
  \hfill
  \begin{subfigure}{0.107\textwidth}
    \includegraphics[width=\linewidth]{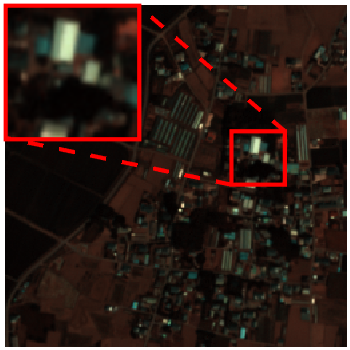}
  \end{subfigure}
  \hfill
  \begin{subfigure}{0.107\textwidth}
    \includegraphics[width=\linewidth]{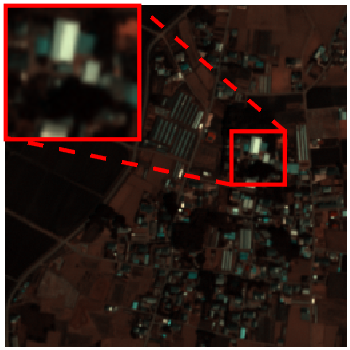}
  \end{subfigure}
  \hfill
  \begin{subfigure}{0.107\textwidth}
    \includegraphics[width=\linewidth]{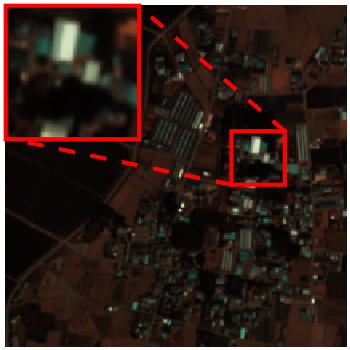}
  \end{subfigure}
  \hfill
  \begin{subfigure}{0.107\textwidth}
    \includegraphics[width=\linewidth]{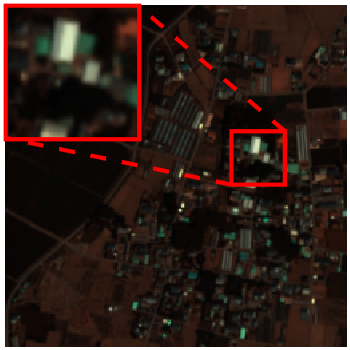}
  \end{subfigure}
  \hfill
  \begin{subfigure}{0.107\textwidth}
    \includegraphics[width=\linewidth]{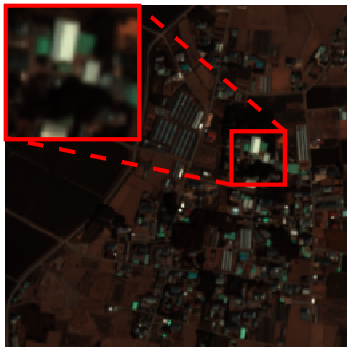}
  \end{subfigure}

  \vspace{0.2mm}

  \begin{subfigure}{0.107\textwidth}
    \includegraphics[width=\linewidth]{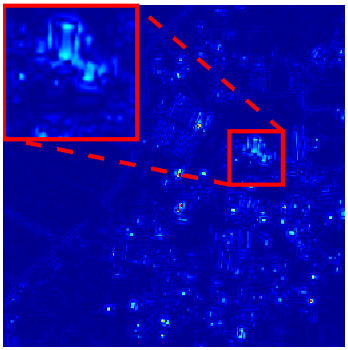}
  \end{subfigure}
  \hfill
  \begin{subfigure}{0.107\textwidth}
    \includegraphics[width=\linewidth]{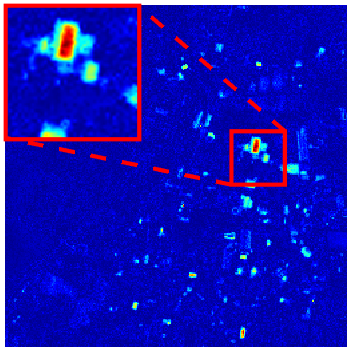}
  \end{subfigure}
  \hfill
  \begin{subfigure}{0.107\textwidth}
    \includegraphics[width=\linewidth]{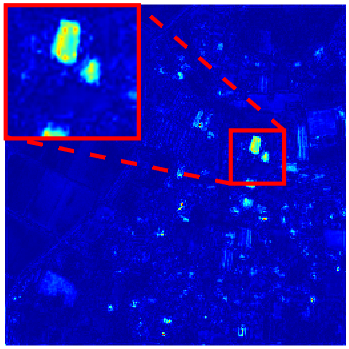}
  \end{subfigure}
  \hfill
  \begin{subfigure}{0.107\textwidth}
    \includegraphics[width=\linewidth]{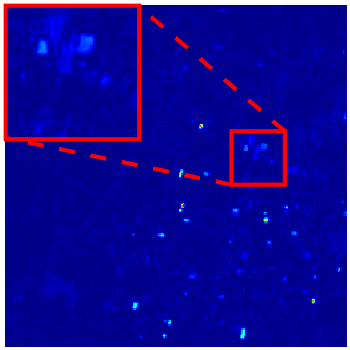}
  \end{subfigure}
  \hfill
  \begin{subfigure}{0.107\textwidth}
    \includegraphics[width=\linewidth]{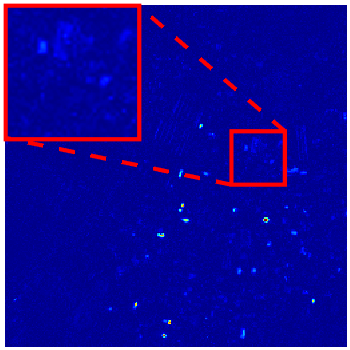}
  \end{subfigure}
  \hfill
  \begin{subfigure}{0.107\textwidth}
    \includegraphics[width=\linewidth]{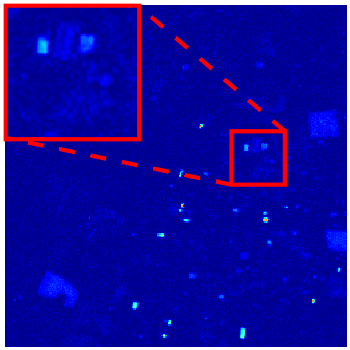}
  \end{subfigure}
  \hfill
  \begin{subfigure}{0.107\textwidth}
    \includegraphics[width=\linewidth]{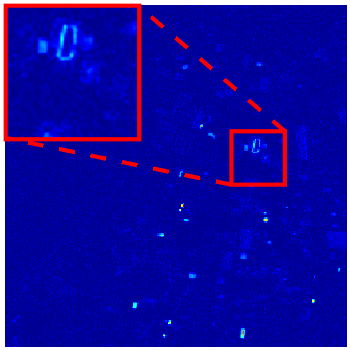}
  \end{subfigure}
  \hfill
  \begin{subfigure}{0.107\textwidth}
    \includegraphics[width=\linewidth]{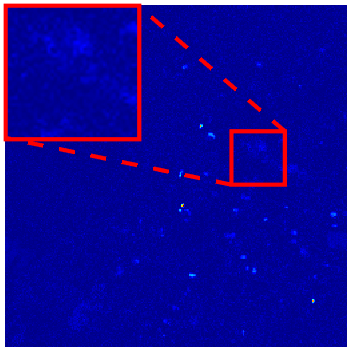}
  \end{subfigure}
  \hfill
  \begin{subfigure}{0.107\textwidth}
    \includegraphics[width=\linewidth]{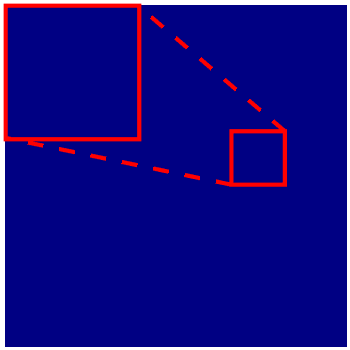}
  \end{subfigure}

  \vspace{0.2mm}
  
  \begin{subfigure}{0.107\textwidth}
    \includegraphics[width=\linewidth]{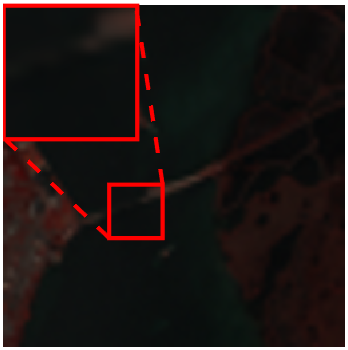}
  \end{subfigure}
  \hfill
  \begin{subfigure}{0.107\textwidth}
    \includegraphics[width=\linewidth]{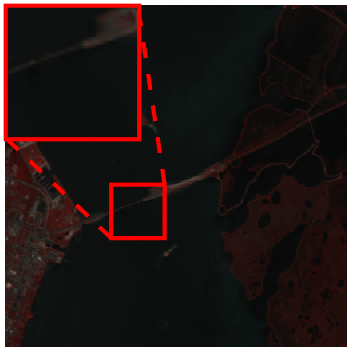}
  \end{subfigure}
  \hfill
  \begin{subfigure}{0.107\textwidth}
    \includegraphics[width=\linewidth]{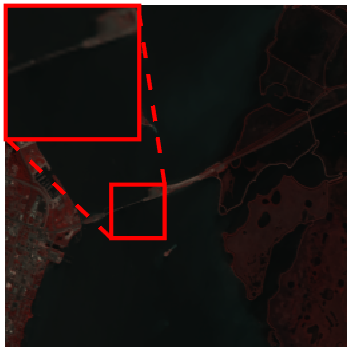}
  \end{subfigure}
  \hfill
  \begin{subfigure}{0.107\textwidth}
    \includegraphics[width=\linewidth]{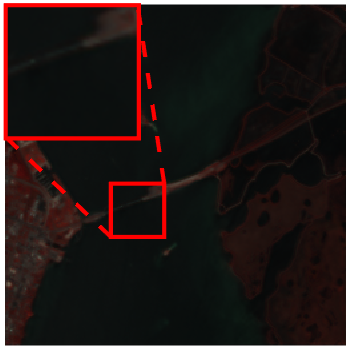}
  \end{subfigure}
  \hfill
  \begin{subfigure}{0.107\textwidth}
    \includegraphics[width=\linewidth]{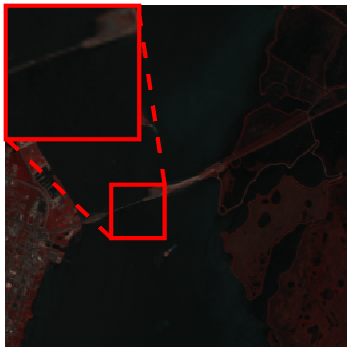}
  \end{subfigure}
  \hfill
  \begin{subfigure}{0.107\textwidth}
    \includegraphics[width=\linewidth]{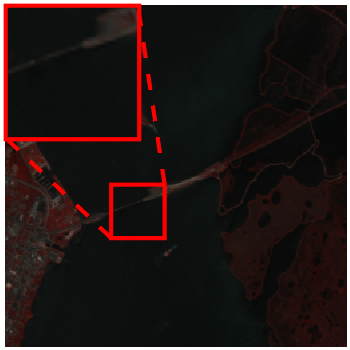}
  \end{subfigure}
  \hfill
  \begin{subfigure}{0.107\textwidth}
    \includegraphics[width=\linewidth]{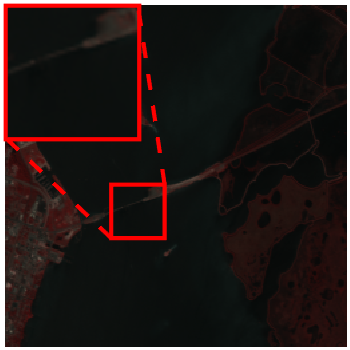}
  \end{subfigure}
  \hfill
  \begin{subfigure}{0.107\textwidth}
    \includegraphics[width=\linewidth]{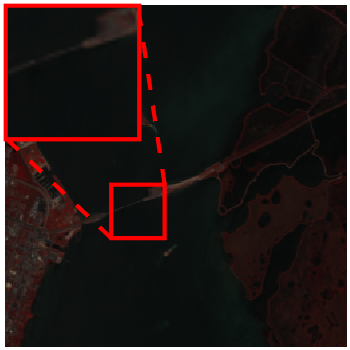}
  \end{subfigure}
  \hfill
  \begin{subfigure}{0.107\textwidth}
    \includegraphics[width=\linewidth]{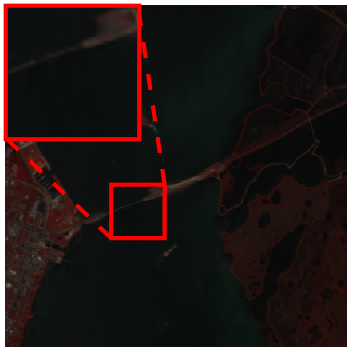}
  \end{subfigure}

  \vspace{0.2mm}

  \begin{subfigure}{0.107\textwidth}
    \captionsetup{font=small}
    \includegraphics[width=\linewidth]{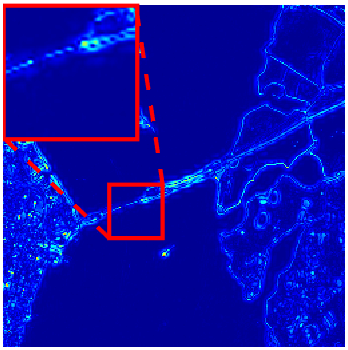}
    \caption*{LR-HSI}
  \end{subfigure}
  \hfill
  \begin{subfigure}{0.107\textwidth}
    \captionsetup{font=small}
    \includegraphics[width=\linewidth]{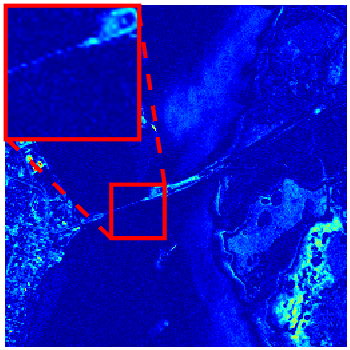}
    \caption*{CNMF}
  \end{subfigure}
  \hfill
  \begin{subfigure}{0.107\textwidth}
    \captionsetup{font=small}
    \includegraphics[width=\linewidth]{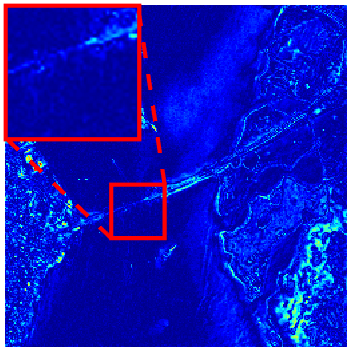}
    \caption*{HySure}
  \end{subfigure}
  \hfill
  \begin{subfigure}{0.107\textwidth}
    \captionsetup{font=small}
    \includegraphics[width=\linewidth]{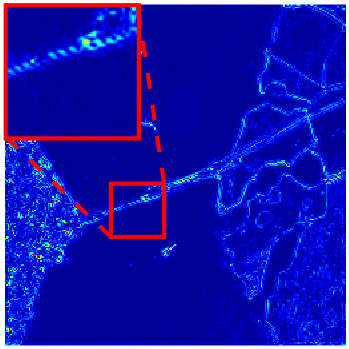}
    \caption*{CuCaNet}
  \end{subfigure}
  \hfill
  \begin{subfigure}{0.107\textwidth}
    \captionsetup{font=small}
    \includegraphics[width=\linewidth]{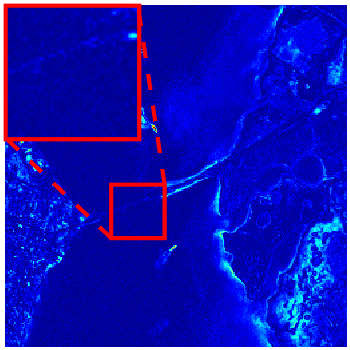}
    \caption*{MIAE}
  \end{subfigure}
  \hfill
  \begin{subfigure}{0.107\textwidth}
    \captionsetup{font=small}
    \includegraphics[width=\linewidth]{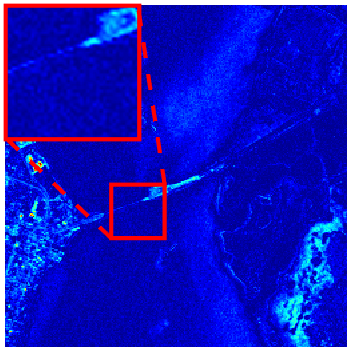}
    \caption*{PLRDiff}
  \end{subfigure}
  \hfill
  \begin{subfigure}{0.107\textwidth}
    \captionsetup{font=small}
     \includegraphics[width=\linewidth]{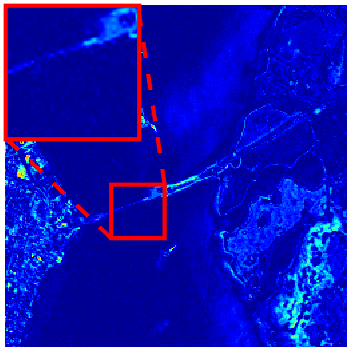}
    \caption*{S$^2$CycleDiff}
  \end{subfigure}
  \hfill
  \begin{subfigure}{0.107\textwidth}
    \captionsetup{font=small}
    \includegraphics[width=\linewidth]{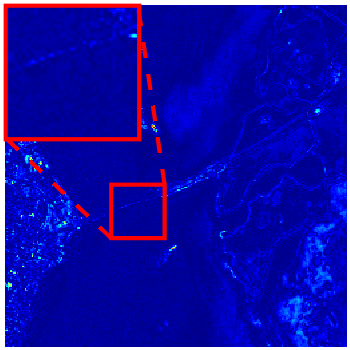}
    \caption*{ARGS-Diff}
  \end{subfigure}
  \hfill
  \begin{subfigure}{0.107\textwidth}
    \captionsetup{font=small}
    \includegraphics[width=\linewidth]{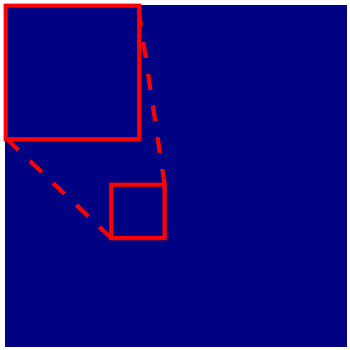}
    \caption*{Reference}
  \end{subfigure}

  \caption{Visual results and reconstruction error maps obtained by different methods. The first, third, and fifth rows are the pseudo-color images of Pavia, Chikusei, and KSC datasets, and the second, fourth, and sixth rows correspond to their error maps. }
  \label{pic:3}
\end{figure*}

Representative visual results are illustrated in Figure \ref{pic:3}, showcasing the superior image quality of proposed method. To further highlight the performance, we include error maps that demonstrate the proposed method’s ability to minimize bias across all three datasets, resulting in high-fidelity HR-HSI reconstructions. Overall, the proposed approach excels in both quantitative and qualitative evaluations, establishing its superiority in HSI-MSI fusion.

\begin{table}[htbp]
    \centering
    \renewcommand{\arraystretch}{1.2}
    \caption{The quantitative results on the Pavia dataset. The \textbf{best} and \underline{second-best} values are highlighted.}
    \label{tab:1} 
    \begin{scriptsize}
    \begin{tabular}{ccc ccc}
        \toprule
        Methods & PSNR\(\uparrow\) & SAM\(\downarrow\) & EGARS\(\downarrow\) & SSIM\(\uparrow\) & Time (s)\\
        \midrule
        CNMF           &38.82&3.13&1.95&0.974&7 \\
        HySure         &37.85&3.19&2.32&0.970&5 \\
        CuCaNet        &38.86&2.95&1.98&0.970&401 \\
        MIAE           &\underline{41.02}&\underline{2.73}&\underline{1.64}&\underline{0.975}&20 \\
        PLRDiff        &40.11&3.40&1.94&0.970&79 \\
        S$^2$CycleDiff &38.77&3.21&2.13&0.970&297 \\
        ARGS-Diff    &\textbf{42.33}&\textbf{2.64}&\textbf{1.49}&\textbf{0.977}&12 \\
        \bottomrule
    \end{tabular}
    \end{scriptsize}
\end{table}

\begin{table}[htbp]
    \centering
    \renewcommand{\arraystretch}{1.2}
    \caption{The quantitative results on the Chikusei dataset. The \textbf{best} and \underline{second-best} values are highlighted.}
    \label{tab:2} 
    \begin{scriptsize}
    \begin{tabular}{ccc ccc}
        \toprule
        Methods & PSNR\(\uparrow\) & SAM\(\downarrow\) & EGARS\(\downarrow\) & SSIM\(\uparrow\) & Time (s)\\
        \midrule
        CNMF           &38.14&2.17&2.59&0.967&11 \\
        HySure         &38.71&2.25&2.65&0.967&5 \\
        CuCaNet        &39.01&2.09&2.55&0.968&433 \\
        MIAE           &\underline{40.53}&\underline{1.95}&\underline{2.51}&\underline{0.969}&24 \\
        PLRDiff        &38.44&2.40&2.79&0.963&79 \\
        S$^2$CycleDiff &39.73&2.08&2.53&0.968&318 \\
        ARGS-Diff    &\textbf{41.90}&\textbf{1.77}&\textbf{2.10}&\textbf{0.970}&12 \\
        \bottomrule
    \end{tabular}
    \end{scriptsize}
\end{table}

\begin{table}[htbp]
    \centering
    \renewcommand{\arraystretch}{1.2}
    \caption{The quantitative results on the KSC dataset. The \textbf{best} and \underline{second-best} values are highlighted.}
    \label{tab:3} 
    \begin{scriptsize}
    \begin{tabular}{ccc ccc}
        \toprule
        Methods & PSNR\(\uparrow\) & SAM\(\downarrow\) & EGARS\(\downarrow\) & SSIM\(\uparrow\) & Time (s)\\
        \midrule
        CNMF          & 39.88  & 3.47 &1.86 & 0.977 & 15       \\
        HySure        & 40.11  & 3.17 &1.80 & 0.978 & 6        \\
        CuCaNet       & 40.39  & 3.05 &1.72 & 0.977 & 474      \\
        MIAE          & \underline{42.36}  & \underline{2.58} &\underline{1.42} & \underline{0.979} & 32       \\
        PLRDiff       & 40.07  & 3.63 &1.83 & 0.977 & 79       \\
        S$^2$CycleDiff& 40.19  & 3.15 &1.79 & 0.978 & 349      \\
        ARGS-Diff     & \textbf{43.63}  & \textbf{2.54} &\textbf{1.25}  & \textbf{0.980} & 13       \\
        \bottomrule
    \end{tabular}
    \end{scriptsize}
\end{table}

\begin{figure*}
  \centering

  \begin{subfigure}{0.107\textwidth}
    \captionsetup{font=small}
    \includegraphics[width=\linewidth]{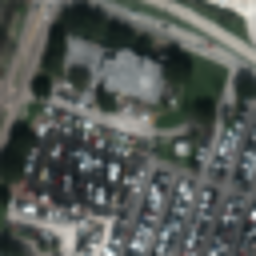}
    \caption*{LR-HSI}
  \end{subfigure}
  \hfill
  \begin{subfigure}{0.107\textwidth}
    \captionsetup{font=small}
    \includegraphics[width=\linewidth]{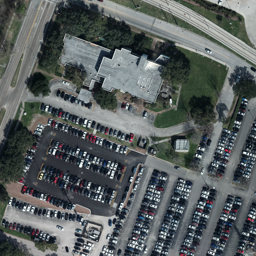}
    \caption*{HR-MSI}
  \end{subfigure}
  \hfill
  \begin{subfigure}{0.107\textwidth}
    \captionsetup{font=small}
    \includegraphics[width=\linewidth]{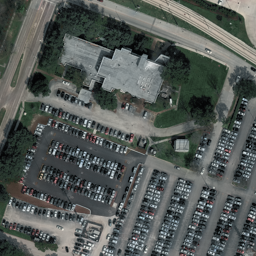}
    \caption*{CNMF}
  \end{subfigure}
  \hfill
  \begin{subfigure}{0.107\textwidth}
    \captionsetup{font=small}
    \includegraphics[width=\linewidth]{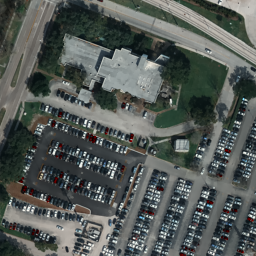}
    \caption*{HySure}
  \end{subfigure}
  \hfill
  \begin{subfigure}{0.107\textwidth}
    \captionsetup{font=small}
    \includegraphics[width=\linewidth]{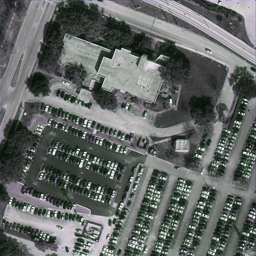}
    \caption*{CuCaNet}
  \end{subfigure}
  \hfill
  \begin{subfigure}{0.107\textwidth}
    \captionsetup{font=small}
    \includegraphics[width=\linewidth]{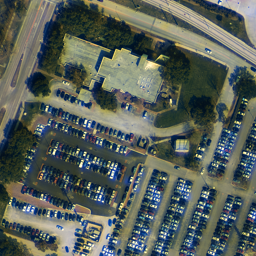}
    \caption*{MIAE}
  \end{subfigure}
  \hfill
  \begin{subfigure}{0.107\textwidth}
    \captionsetup{font=small}
     \includegraphics[width=\linewidth]{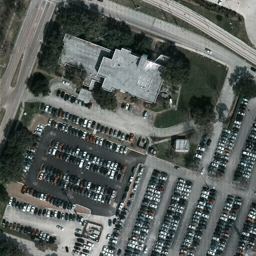}
    \caption*{PLRDiff}
  \end{subfigure}
  \hfill
  \begin{subfigure}{0.107\textwidth}
    \captionsetup{font=small}
    \includegraphics[width=\linewidth]{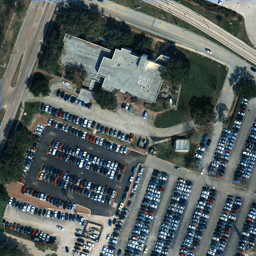}
    \caption*{ARGS-Diff}
  \end{subfigure}

  \caption{Visual results obtained by different methods on the Houston dataset.}
  \label{pic:real}
\end{figure*}

\subsubsection{Results on Real-Data}

Figure \ref{pic:real} presents the reconstructed pseudo-color images from the DFC2018 Houston dataset, generated using various methods. The reconstructed HSIs from the comparison methods exhibit varying degrees of deficiencies in color consistency, naturalness, and saturation control. In contrast, ARGS-Diff effectively preserves the authenticity and vibrancy of colors while avoiding pseudo-color artifacts and color distortions. The resulting spectral images exhibit high quality, closely resembling real-world scenes. This demonstrates ARGS-Diff's superior ability to preserve fine details and maintain overall color balance, ensuring the consistency of both spatial and spectral information in the reconstructed hyperspectral images. These results highlight the effectiveness of the proposed method, particularly in real-world applications, where accurate color reproduction and detail preservation are critical.

\subsection{Ablation Study}


In this section, we assess the impact of the Adaptive Residual Guided Module (ARGM) on the performance of the proposed method. The results, summarized in Table \ref{tab:4}, clearly demonstrate that incorporating ARGM leads to a significant improvement in the quality of the reconstructed images. Specifically, the inclusion of ARGM increased the PSNR by 0.47 dB, 0.57 dB, and 0.61 dB across the three datasets, while the SAM improved by 0.11, 0.12, and 0.16, underscoring its positive contribution to the fusion performance. Importantly, the computational overhead introduced by ARGM is minimal, adding only 1-2 seconds to the overall runtime, which has a negligible impact on the total processing time. 

\begin{figure}[htbp]
  \centering

  \begin{subfigure}{0.075\textwidth}
    \includegraphics[width=\linewidth]{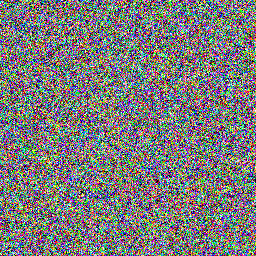}
  \end{subfigure}
  \hfill
  \begin{subfigure}{0.075\textwidth}
    \includegraphics[width=\linewidth]{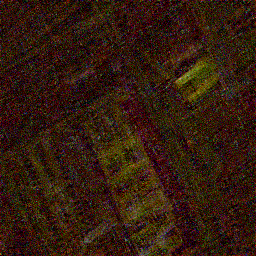}
  \end{subfigure}
  \hfill
  \begin{subfigure}{0.075\textwidth}
    \includegraphics[width=\linewidth]{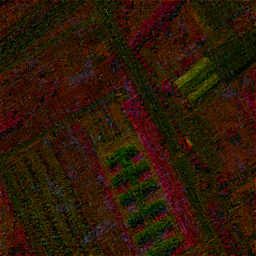}
  \end{subfigure}
  \hfill
  \begin{subfigure}{0.075\textwidth}
    \includegraphics[width=\linewidth]{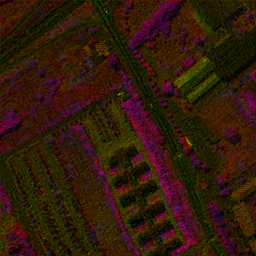}
  \end{subfigure}
  \hfill
  \begin{subfigure}{0.075\textwidth}
    \includegraphics[width=\linewidth]{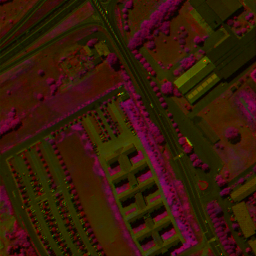}
  \end{subfigure}
  \hfill
  \begin{subfigure}{0.075\textwidth}
    \includegraphics[width=\linewidth]{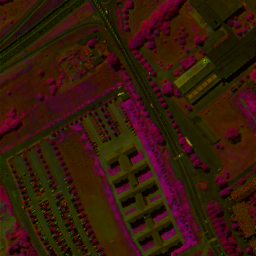}
  \end{subfigure}

  \vspace{0.2mm}

  \begin{subfigure}{0.075\textwidth}
    \captionsetup{font=scriptsize}
    \includegraphics[width=\linewidth]{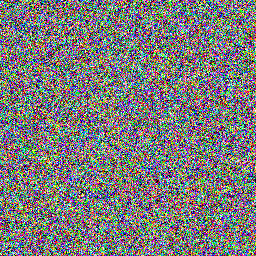}
    \caption*{$t=500$}
  \end{subfigure}
  \hfill
  \begin{subfigure}{0.075\textwidth}
    \captionsetup{font=scriptsize}
    \includegraphics[width=\linewidth]{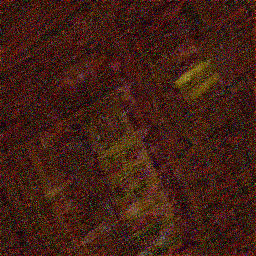}
    \caption*{$t=400$}
  \end{subfigure}
  \hfill
  \begin{subfigure}{0.075\textwidth}
    \captionsetup{font=scriptsize}
    \includegraphics[width=\linewidth]{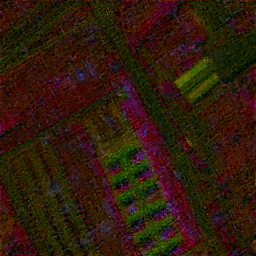}
    \caption*{$t=300$}
  \end{subfigure}
  \hfill
  \begin{subfigure}{0.075\textwidth}
    \captionsetup{font=scriptsize}
     \includegraphics[width=\linewidth]{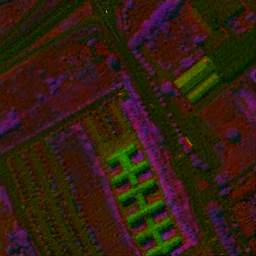}
    \caption*{$t=200$}
  \end{subfigure}
  \hfill
  \begin{subfigure}{0.075\textwidth}
    \captionsetup{font=scriptsize}
    \includegraphics[width=\linewidth]{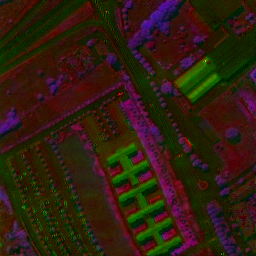}
    \caption*{$t=100$}
  \end{subfigure}
  \hfill
  \begin{subfigure}{0.075\textwidth}
    \captionsetup{font=scriptsize}
    \includegraphics[width=\linewidth]{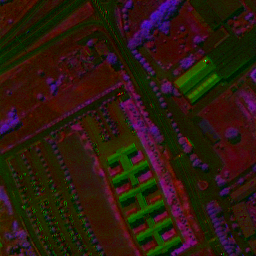}
    \caption*{$t=0$}
  \end{subfigure}

  \caption{Intermediate results of the reduced coefficient $\mathcal{A}$. The first and second rows represent 'w/o ARGM' and 'w/ ARGM'.}
  \label{pic:AGRM}
\end{figure}

Additionally, Figure \ref{pic:AGRM} presents the reconstruction results of the reduced coefficient $\mathcal{A}$ at different sampling steps \( t = \{500, 400, 300, 200, 100, 0\} \) on the Pavia dataset. It is evident that, with the incorporation of ARGM, the reconstructed reduced coefficient becomes more informative and detailed. This visual improvement further confirms the significant enhancement brought by ARGM, which facilitates more accurate and robust fusion, contributing to better alignment of the spectral and spatial components.

\begin{table}[htbp]
    \centering
    \renewcommand{\arraystretch}{1.2}
    \caption{The quantitative ablation results of ARGM.}
    \label{tab:4}  
    \begin{scriptsize}
    \begin{tabular}{ccccc}
        \toprule
        Datasets & Methods & PSNR & SAM & Time (s)\\
        \midrule
        \multirow{2}{*}{Pavia}
        &w/o ARGM &41.86&2.75&11\\
        &w/ ARGM  &42.33&2.64&12\\
        \midrule
        \multirow{2}{*}{Chikusei}
        &w/o ARGM &41.33&1.89&11\\
        &w/ ARGM  &41.90&1.77&12\\
        \midrule
        \multirow{2}{*}{KSC}
        &w/o ARGM &43.02&2.70&11\\
        &w/ ARGM  &43.63&2.54&13\\
        \bottomrule
    \end{tabular}
    \end{scriptsize}
\end{table}

\subsection{Analysis of Hyperparameters}

\begin{figure}[htbp]
  \centering
  
  \begin{subfigure}{0.235\textwidth}
    \includegraphics[width=\linewidth]{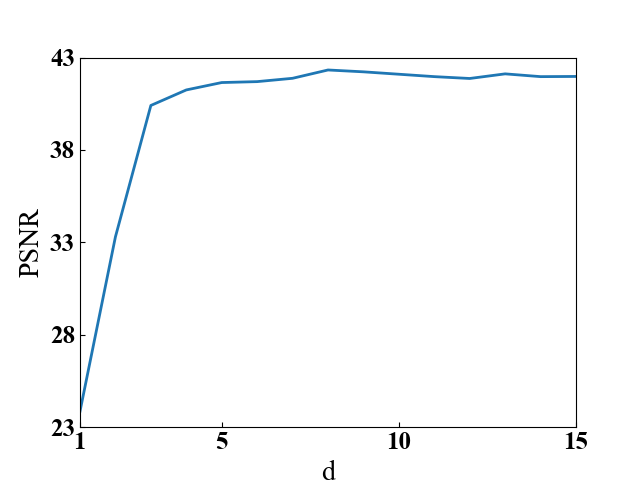}
  \end{subfigure}
  \hspace{-0.025\textwidth}
  \begin{subfigure}{0.235\textwidth}
    \includegraphics[width=\linewidth]{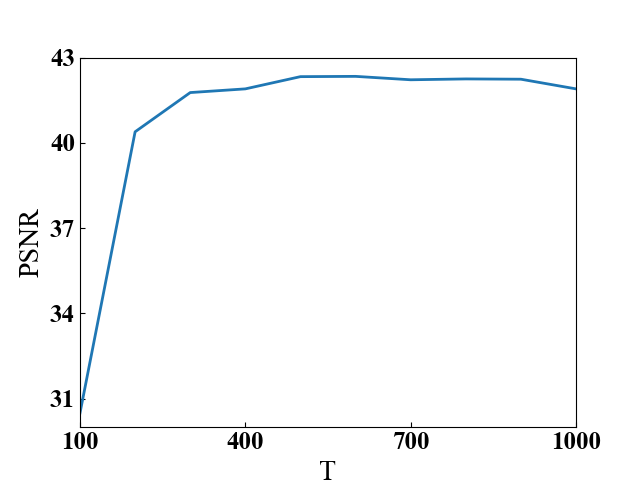}
  \end{subfigure}
  
  \caption{Sensitivity analysis of the parameters $d$ and $T$.}
  \label{fig:RT}
\end{figure}

$d$ \textbf{and} $T$. In this part, we examine the parameter sensitivity related to the number of subspace dimensions and the number of sampling steps. Using the Pavia dataset as a case study, the results are shown in Figure \ref{fig:RT}. As \( d \) increases from 1 to 8, performance improves significantly; however, beyond this point, performance gradually declines. This drop can be attributed to the increasing number of parameters that need to be updated when d becomes too large, which complicates the convergence process. Consequently, we set \( d \) to 8. 
For the number of sampling steps \( T \), we observe substantial improvement as \( T \) increases from 100 to 500. However, beyond 500 steps, the performance reaches a plateau and may even slightly decrease, suggesting that the model has reached saturation and further increases may lead to overfitting. Therefore, we set \( T \) to 500.

\noindent $\rho_1$ \textbf{and} $\rho_2$. In Eq. (\ref{f:9}), we introduce two hyperparameters, $\rho_1$ and $\rho_2$, which control the update step sizes for the predicted components. To determine an appropriate combination of values, we conduct a hyperparameter search over $\rho_1, \rho_2 \in \{0.01, 0.02, 0.03, 0.04, 0.05, 0.06, 0.07\}$ using the Pavia dataset. The PSNR results are shown in Table \ref{tab:rho}, from which we observe that the best performance is achieved when $\rho_1 = \rho_2 = 0.05$. Therefore, for the Pavia dataset, we set $\rho_1$ and $\rho_2$ to 0.05.

\setlength{\heavyrulewidth}{0.9pt}

\begin{table}[h]
\centering
\renewcommand{\arraystretch}{1.5}
\caption{Sensitivity analysis of the parameters $\rho_1$ and $\rho_2$ on Pavia dataset. The \textbf{best} result is highlighted.}
\label{tab:rho}
\begin{scriptsize}
\begin{tabularx}{0.47\textwidth}{c|ccccccc}
\specialrule{\heavyrulewidth}{0pt}{0pt}
\backslashbox{$\rho_1$}{$\rho_2$} & 0.01   & 0.02  & 0.03   & 0.04  & 0.05   & 0.06  & 0.07 \\
\hline
0.01    & 33.70 & 38.96 & 34.09 & 36.88 & 34.67 & 35.07 & 31.84 \\
0.02    & 40.75 & 40.69 & 40.82 & 41.15 & 41.41 & 41.50 & 40.73 \\
0.03    & 41.35 & 41.55 & 41.71 & 41.57 & 41.59 & 41.91 & 41.80 \\
0.04    & 41.75 & 41.89 & 41.77 & 41.93 & 41.90 & 42.10 & 41.69 \\
0.05    & 41.72 & 42.00 & 42.13 & 42.25 & \textbf{42.33} & 42.27 & 41.67 \\
0.06    & 41.66 & 41.97 & 42.13 & 42.15 & 42.23 & 42.11 & 41.95 \\
0.07    & 41.67 & 41.80 & 41.97 & 42.19 & 42.14 & 41.78 & 41.94 \\
\specialrule{\heavyrulewidth}{0pt}{0pt}
\end{tabularx}
\end{scriptsize}
\end{table}



\subsection{Inference Time and Model Scale Anaiysis}

In this section, we compare the proposed method with other diffusion model approaches in terms of parameter count, memory consumption, and runtime. Table \ref{tab:time} summarizes the results for the PLRDiff, S$^2$CycleDiff, and the proposed ARGS-Diff, using the Pavia dataset as an example. The proposed method has a parameter count of just 21.85M, which is approximately 1/20th and 1/30th of the parameter counts for PLRDiff and S$^2$CycleDiff, respectively. It also consumes only 2.11GB of memory, significantly less than the other two methods. Furthermore, the proposed method completes sampling in just 12 seconds, significantly outperforming PLRDiff (79 seconds) and S$^2$CycleDiff (297 seconds) in terms of processing time. Overall, the proposed method not only excels in fusion accuracy but also offers substantial advantages in terms of model size, memory usage, and runtime, making it highly suitable for real-world applications.

\begin{table}[htbp]
    \centering
    \renewcommand{\arraystretch}{1.2}
    \caption{The model size and computational consumption of the proposed method and other diffusion model-based methdods.}
    \label{tab:time}  
    \begin{scriptsize}  
    \begin{tabular}{ccc ccc}
        \toprule
        Methods & PSNR & Params (M) & Memory (GB) & Time (s) \\
        \midrule
        PLRDiff        &40.11&391.05&6.21 &79 \\
        S$^2$CycleDiff &38.77&606.44&23.34&297 \\
        ARGS-Diff    &42.33&21.85 &2.11 &12 \\
        \bottomrule
    \end{tabular}
    \end{scriptsize}
\end{table}

\section{Conclusion}
\label{sec:conclusion}

In this paper, we propose a self-learning Adaptive Residual Guided Subspace Diffusion Model (ARGS-Diff) for HSI-MSI fusion. We carefully design two lightweight spectral and spatial networks, training them using the observed LR-HSI and HR-MSI to extract the rich spectral and spatial information inherent in each. These well-trained networks are then employed to reconstruct two low-dimensional components: the spectral basis and the reduced coefficient. The HR-HSI is obtained by taking the product of these two components. Additionally, we introduce the Adaptive Residual Guided Module (ARGM), which further enhances the sampling process by stabilizing the updates of the spectral and spatial components. Extensive experimental evaluations demonstrate the superiority of the proposed method, achieving remarkable performance in the field of HSI-MSI fusion.



{
    \small
    \bibliographystyle{ieeenat_fullname}
    \bibliography{main}

\begin{thebibliography}{42}
\providecommand{\natexlab}[1]{#1}
\providecommand{\url}[1]{\texttt{#1}}
\expandafter\ifx\csname urlstyle\endcsname\relax
  \providecommand{\doi}[1]{doi: #1}\else
  \providecommand{\doi}{doi: \begingroup \urlstyle{rm}\Url}\fi

\bibitem[Bansal et~al.(2023)Bansal, Chu, Schwarzschild, Sengupta, Goldblum, Geiping, and Goldstein]{bansal2023universal}
Arpit Bansal, Hong-Min Chu, Avi Schwarzschild, Soumyadip Sengupta, Micah Goldblum, Jonas Geiping, and Tom Goldstein.
\newblock Universal guidance for diffusion models.
\newblock In \emph{Proceedings of the IEEE/CVF Conference on Computer Vision and Pattern Recognition}, pages 843--852, 2023.

\bibitem[Chung et~al.(2023)Chung, Kim, Mccann, Klasky, and Ye]{chung2022diffusion}
Hyungjin Chung, Jeongsol Kim, Michael~Thompson Mccann, Marc~Louis Klasky, and Jong~Chul Ye.
\newblock Diffusion posterior sampling for general noisy inverse problems.
\newblock In \emph{The Eleventh International Conference on Learning Representations}, 2023.

\bibitem[Cutler and Breiman(1994)]{cutler1994archetypal}
Adele Cutler and Leo Breiman.
\newblock Archetypal analysis.
\newblock \emph{Technometrics}, 36\penalty0 (4):\penalty0 338--347, 1994.

\bibitem[Dale et~al.(2013)Dale, Thewis, Boudry, Rotar, Dardenne, Baeten, and Pierna]{dale2013hyperspectral}
Laura~M Dale, Andr{\'e} Thewis, Christelle Boudry, Ioan Rotar, Pierre Dardenne, Vincent Baeten, and Juan A~Fern{\'a}ndez Pierna.
\newblock Hyperspectral imaging applications in agriculture and agro-food product quality and safety control: A review.
\newblock \emph{Applied Spectroscopy Reviews}, 48\penalty0 (2):\penalty0 142--159, 2013.

\bibitem[Dhariwal and Nichol(2021)]{dhariwal2021diffusion}
Prafulla Dhariwal and Alexander Nichol.
\newblock Diffusion models beat gans on image synthesis.
\newblock \emph{Advances in neural information processing systems}, 34:\penalty0 8780--8794, 2021.

\bibitem[Dosovitskiy(2020)]{dosovitskiy2020image}
Alexey Dosovitskiy.
\newblock An image is worth 16x16 words: Transformers for image recognition at scale.
\newblock \emph{arXiv preprint arXiv:2010.11929}, 2020.

\bibitem[Guo et~al.(2023)Guo, Xie, Jiang, Li, Lei, and Fang]{guo2023toward}
Wen-jin Guo, Weiying Xie, Kai Jiang, Yunsong Li, Jie Lei, and Leyuan Fang.
\newblock Toward stable, interpretable, and lightweight hyperspectral super-resolution.
\newblock In \emph{Proceedings of the IEEE/CVF Conference on Computer Vision and Pattern Recognition}, pages 22272--22281, 2023.

\bibitem[Ho et~al.(2020)Ho, Jain, and Abbeel]{ho2020denoising}
Jonathan Ho, Ajay Jain, and Pieter Abbeel.
\newblock Denoising diffusion probabilistic models.
\newblock \emph{Advances in neural information processing systems}, 33:\penalty0 6840--6851, 2020.

\bibitem[Huang et~al.(2017)Huang, Wen, Li, and Qin]{huang2017multi}
Xin Huang, Dawei Wen, Jiayi Li, and Rongjun Qin.
\newblock Multi-level monitoring of subtle urban changes for the megacities of china using high-resolution multi-view satellite imagery.
\newblock \emph{Remote Sensing of Environment}, 196:\penalty0 56--75, 2017.

\bibitem[Khanal et~al.(2020)Khanal, Kc, Fulton, Shearer, and Ozkan]{khanal2020remote}
Sami Khanal, Kushal Kc, John~P Fulton, Scott Shearer, and Erdal Ozkan.
\newblock Remote sensing in agriculture—accomplishments, limitations, and opportunities.
\newblock \emph{Remote Sensing}, 12\penalty0 (22):\penalty0 3783, 2020.

\bibitem[Kingma(2014)]{kingma2014adam}
Diederik~P Kingma.
\newblock Adam: A method for stochastic optimization.
\newblock \emph{arXiv preprint arXiv:1412.6980}, 2014.

\bibitem[Li et~al.(2020)Li, Pei, Zhao, Xiao, Sang, and Zhang]{li2020review}
Jun Li, Yanqiu Pei, Shaohua Zhao, Rulin Xiao, Xiao Sang, and Chengye Zhang.
\newblock A review of remote sensing for environmental monitoring in china.
\newblock \emph{Remote Sensing}, 12\penalty0 (7):\penalty0 1130, 2020.

\bibitem[Liu et~al.(2022)Liu, Wu, Xiao, and Wu]{liu2022model}
Jianjun Liu, Zebin Wu, Liang Xiao, and Xiao-Jun Wu.
\newblock Model inspired autoencoder for unsupervised hyperspectral image super-resolution.
\newblock \emph{IEEE Transactions on Geoscience and Remote Sensing}, 60:\penalty0 1--12, 2022.

\bibitem[Liu et~al.(2013)Liu, Zhou, Jiang, Zhuang, Mansaray, and Zhang]{liu2013targeting}
Lei Liu, Jun Zhou, Dong Jiang, Dafang Zhuang, Lamin~R Mansaray, and Bing Zhang.
\newblock Targeting mineral resources with remote sensing and field data in the xiemisitai area, west junggar, xinjiang, china.
\newblock \emph{Remote sensing}, 5\penalty0 (7):\penalty0 3156--3171, 2013.

\bibitem[Liu et~al.(2021)Liu, Lin, Cao, Hu, Wei, Zhang, Lin, and Guo]{liu2021swin}
Ze Liu, Yutong Lin, Yue Cao, Han Hu, Yixuan Wei, Zheng Zhang, Stephen Lin, and Baining Guo.
\newblock Swin transformer: Hierarchical vision transformer using shifted windows.
\newblock In \emph{Proceedings of the IEEE/CVF international conference on computer vision}, pages 10012--10022, 2021.

\bibitem[Loncan et~al.(2015)Loncan, De~Almeida, Bioucas-Dias, Briottet, Chanussot, Dobigeon, Fabre, Liao, Licciardi, Simoes, et~al.]{loncan2015hyperspectral}
Laetitia Loncan, Luis~B De~Almeida, Jos{\'e}~M Bioucas-Dias, Xavier Briottet, Jocelyn Chanussot, Nicolas Dobigeon, Sophie Fabre, Wenzhi Liao, Giorgio~A Licciardi, Miguel Simoes, et~al.
\newblock Hyperspectral pansharpening: A review.
\newblock \emph{IEEE Geoscience and Remote Sensing Magazine}, 3\penalty0 (3):\penalty0 27--46, 2015.

\bibitem[Lu et~al.(2022)Lu, Zhou, Bao, Chen, LI, and Zhu]{lu2022fast}
Cheng Lu, Yuhao Zhou, Fan Bao, Jianfei Chen, Chongxuan LI, and Jun Zhu.
\newblock Dpm-solver: A fast ode solver for diffusion probabilistic model sampling in around 10 steps.
\newblock In \emph{Advances in Neural Information Processing Systems}, pages 5775--5787, 2022.

\bibitem[Miao et~al.(2023)Miao, Zhang, Zhang, and Tao]{miao2023dds2m}
Yuchun Miao, Lefei Zhang, Liangpei Zhang, and Dacheng Tao.
\newblock Dds2m: Self-supervised denoising diffusion spatio-spectral model for hyperspectral image restoration.
\newblock In \emph{Proceedings of the IEEE/CVF International Conference on Computer Vision}, pages 12086--12096, 2023.

\bibitem[Nichol and Dhariwal(2021)]{nichol2021improved}
Alexander~Quinn Nichol and Prafulla Dhariwal.
\newblock Improved denoising diffusion probabilistic models.
\newblock In \emph{International conference on machine learning}, pages 8162--8171. PMLR, 2021.

\bibitem[Pang et~al.(2024)Pang, Rui, Cui, Wang, Meng, and Cao]{pang2024hir}
Li Pang, Xiangyu Rui, Long Cui, Hongzhong Wang, Deyu Meng, and Xiangyong Cao.
\newblock Hir-diff: Unsupervised hyperspectral image restoration via improved diffusion models.
\newblock In \emph{Proceedings of the IEEE/CVF Conference on Computer Vision and Pattern Recognition}, pages 3005--3014, 2024.

\bibitem[Qu et~al.(2024)Qu, He, Dong, and Zhao]{qu2024s2cyclediff}
Jiahui Qu, Jie He, Wenqian Dong, and Jingyu Zhao.
\newblock S2cyclediff: Spatial-spectral-bilateral cycle-diffusion framework for hyperspectral image super-resolution.
\newblock In \emph{Proceedings of the AAAI Conference on Artificial Intelligence}, pages 4623--4631, 2024.

\bibitem[Ronneberger et~al.(2015)Ronneberger, Fischer, and Brox]{ronneberger2015u}
Olaf Ronneberger, Philipp Fischer, and Thomas Brox.
\newblock U-net: Convolutional networks for biomedical image segmentation.
\newblock In \emph{Medical image computing and computer-assisted intervention--MICCAI 2015: 18th international conference, Munich, Germany, October 5-9, 2015, proceedings, part III 18}, pages 234--241. Springer, 2015.

\bibitem[Rui et~al.(2024)Rui, Cao, Pang, Zhu, Yue, and Meng]{rui2023unsupervised}
Xiangyu Rui, Xiangyong Cao, Li Pang, Zeyu Zhu, Zongsheng Yue, and Deyu Meng.
\newblock Unsupervised hyperspectral pansharpening via low-rank diffusion model.
\newblock \emph{Information Fusion}, 107:\penalty0 102325, 2024.

\bibitem[Saharia et~al.(2022)Saharia, Ho, Chan, Salimans, Fleet, and Norouzi]{saharia2022image}
Chitwan Saharia, Jonathan Ho, William Chan, Tim Salimans, David~J Fleet, and Mohammad Norouzi.
\newblock Image super-resolution via iterative refinement.
\newblock \emph{IEEE Transactions on Pattern Analysis and Machine Intelligence}, 45\penalty0 (4):\penalty0 4713--4726, 2022.

\bibitem[Scarpa et~al.(2018)Scarpa, Vitale, and Cozzolino]{scarpa2018target}
Giuseppe Scarpa, Sergio Vitale, and Davide Cozzolino.
\newblock Target-adaptive cnn-based pansharpening.
\newblock \emph{IEEE Transactions on Geoscience and Remote Sensing}, 56\penalty0 (9):\penalty0 5443--5457, 2018.

\bibitem[Shah et~al.(2008)Shah, Younan, and King]{shah2008efficient}
Vijay~P Shah, Nicolas~H Younan, and Roger~L King.
\newblock An efficient pan-sharpening method via a combined adaptive pca approach and contourlets.
\newblock \emph{IEEE Transactions on Geoscience and Remote Sensing}, 46\penalty0 (5):\penalty0 1323--1335, 2008.

\bibitem[Shirmard et~al.(2022)Shirmard, Farahbakhsh, M{\"u}ller, and Chandra]{shirmard2022review}
Hojat Shirmard, Ehsan Farahbakhsh, R~Dietmar M{\"u}ller, and Rohitash Chandra.
\newblock A review of machine learning in processing remote sensing data for mineral exploration.
\newblock \emph{Remote Sensing of Environment}, 268:\penalty0 112750, 2022.

\bibitem[Simoes et~al.(2014)Simoes, Bioucas-Dias, Almeida, and Chanussot]{simoes2014convex}
Miguel Simoes, Jos{\'e} Bioucas-Dias, Luis~B Almeida, and Jocelyn Chanussot.
\newblock A convex formulation for hyperspectral image superresolution via subspace-based regularization.
\newblock \emph{IEEE Transactions on Geoscience and Remote Sensing}, 53\penalty0 (6):\penalty0 3373--3388, 2014.

\bibitem[Song et~al.(2021)Song, Meng, and Ermon]{song2020denoising}
Jiaming Song, Chenlin Meng, and Stefano Ermon.
\newblock Denoising diffusion implicit models.
\newblock In \emph{International Conference on Learning Representations}, 2021.

\bibitem[Song et~al.(2020)Song, Sohl-Dickstein, Kingma, Kumar, Ermon, and Poole]{song2020score}
Yang Song, Jascha Sohl-Dickstein, Diederik~P Kingma, Abhishek Kumar, Stefano Ermon, and Ben Poole.
\newblock Score-based generative modeling through stochastic differential equations.
\newblock \emph{arXiv preprint arXiv:2011.13456}, 2020.

\bibitem[Vaswani et~al.(2017)Vaswani, Shazeer, Parmar, Uszkoreit, Jones, Gomez, Kaiser, and Polosukhin]{vaswani2017attention}
Ashish Vaswani, Noam Shazeer, Niki Parmar, Jakob Uszkoreit, Llion Jones, Aidan~N Gomez, \L~ukasz Kaiser, and Illia Polosukhin.
\newblock Attention is all you need.
\newblock In \emph{Advances in Neural Information Processing Systems}, 2017.

\bibitem[Wang et~al.(2019)Wang, Zeng, Huang, Ding, and Paisley]{wang2019deep}
Wu Wang, Weihong Zeng, Yue Huang, Xinghao Ding, and John Paisley.
\newblock Deep blind hyperspectral image fusion.
\newblock In \emph{Proceedings of the IEEE/CVF International Conference on Computer Vision}, pages 4150--4159, 2019.

\bibitem[Wang et~al.(2021)Wang, Xie, Li, Fan, Song, Liang, Lu, Luo, and Shao]{wang2021pyramid}
Wenhai Wang, Enze Xie, Xiang Li, Deng-Ping Fan, Kaitao Song, Ding Liang, Tong Lu, Ping Luo, and Ling Shao.
\newblock Pyramid vision transformer: A versatile backbone for dense prediction without convolutions.
\newblock In \emph{Proceedings of the IEEE/CVF international conference on computer vision}, pages 568--578, 2021.

\bibitem[Wang et~al.(2023)Wang, Yu, and Zhang]{wang2022zero}
Yinhuai Wang, Jiwen Yu, and Jian Zhang.
\newblock Zero-shot image restoration using denoising diffusion null-space model.
\newblock In \emph{The Eleventh International Conference on Learning Representations}, 2023.

\bibitem[Wu et~al.(2023)Wu, Wang, Bai, Mao, Li, and Shen]{wu2023hsr}
Chanyue Wu, Dong Wang, Yunpeng Bai, Hanyu Mao, Ying Li, and Qiang Shen.
\newblock Hsr-diff: Hyperspectral image super-resolution via conditional diffusion models.
\newblock In \emph{Proceedings of the IEEE/CVF International Conference on Computer Vision}, pages 7083--7093, 2023.

\bibitem[Xie et~al.(2020)Xie, Zhou, Zhao, Xu, and Meng]{xie2020mhf}
Qi Xie, Minghao Zhou, Qian Zhao, Zongben Xu, and Deyu Meng.
\newblock Mhf-net: An interpretable deep network for multispectral and hyperspectral image fusion.
\newblock \emph{IEEE Transactions on Pattern Analysis and Machine Intelligence}, 44\penalty0 (3):\penalty0 1457--1473, 2020.

\bibitem[Xue et~al.(2022)Xue, Ye, Hu, Zhu, and Wang]{xue2022ddrm}
Qiao Xue, Qingqing Ye, Haibo Hu, Youwen Zhu, and Jian Wang.
\newblock Ddrm: A continual frequency estimation mechanism with local differential privacy.
\newblock \emph{IEEE Transactions on Knowledge and Data Engineering}, 35\penalty0 (7):\penalty0 6784--6797, 2022.

\bibitem[Yang et~al.(2018)Yang, Zhao, and Chan]{yang2018hyperspectral}
Jingxiang Yang, Yong-Qiang Zhao, and Jonathan Cheung-Wai Chan.
\newblock Hyperspectral and multispectral image fusion via deep two-branches convolutional neural network.
\newblock \emph{Remote Sensing}, 10\penalty0 (5):\penalty0 800, 2018.

\bibitem[Yao et~al.(2020)Yao, Hong, Chanussot, Meng, Zhu, and Xu]{yao2020cross}
Jing Yao, Danfeng Hong, Jocelyn Chanussot, Deyu Meng, Xiaoxiang Zhu, and Zongben Xu.
\newblock Cross-attention in coupled unmixing nets for unsupervised hyperspectral super-resolution.
\newblock In \emph{Computer Vision -- ECCV 2020}, pages 208--224, 2020.

\bibitem[Yokoya et~al.(2011)Yokoya, Yairi, and Iwasaki]{yokoya2011coupled}
Naoto Yokoya, Takehisa Yairi, and Akira Iwasaki.
\newblock Coupled nonnegative matrix factorization unmixing for hyperspectral and multispectral data fusion.
\newblock \emph{IEEE Transactions on Geoscience and Remote Sensing}, 50\penalty0 (2):\penalty0 528--537, 2011.

\bibitem[Zhang et~al.(2020)Zhang, Huang, Wang, and Li]{zhang2020ssr}
Xueting Zhang, Wei Huang, Qi Wang, and Xuelong Li.
\newblock Ssr-net: Spatial--spectral reconstruction network for hyperspectral and multispectral image fusion.
\newblock \emph{IEEE Transactions on Geoscience and Remote Sensing}, 59\penalty0 (7):\penalty0 5953--5965, 2020.

\bibitem[Zhou et~al.(2024)Zhou, Huang, Yan, Hong, Jia, Chanussot, and Li]{zhou2024general}
Man Zhou, Jie Huang, Keyu Yan, Danfeng Hong, Xiuping Jia, Jocelyn Chanussot, and Chongyi Li.
\newblock A general spatial-frequency learning framework for multimodal image fusion.
\newblock \emph{IEEE Transactions on Pattern Analysis and Machine Intelligence}, 2024.

\end{thebibliography}
}

\clearpage
\setcounter{page}{1}
\maketitlesupplementary

\section{Details of Network Architecture}
\label{sec:na}

\begin{figure}[htbp]
  \vspace{-5mm}
  \centering
  
  \begin{subfigure}{0.5\textwidth}
    \captionsetup{font=small}
    \hspace{-7mm}
    \includegraphics[width=1.2\linewidth]{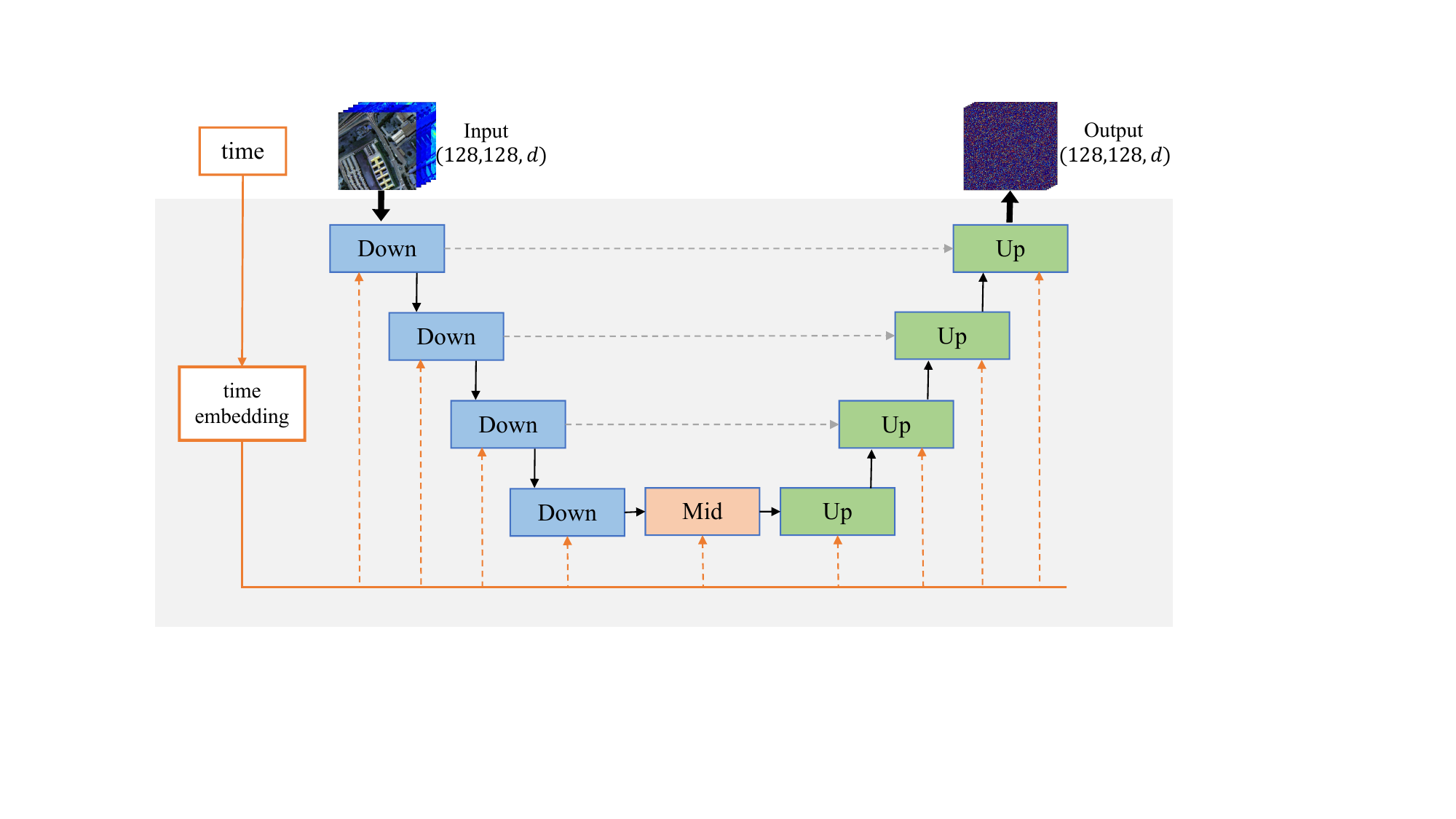}
    \vspace{-15mm}
    \caption{Spatial Network}
  \end{subfigure}
  \begin{subfigure}{0.5\textwidth}
    \captionsetup{font=small}
    \includegraphics[width=1.2\linewidth]{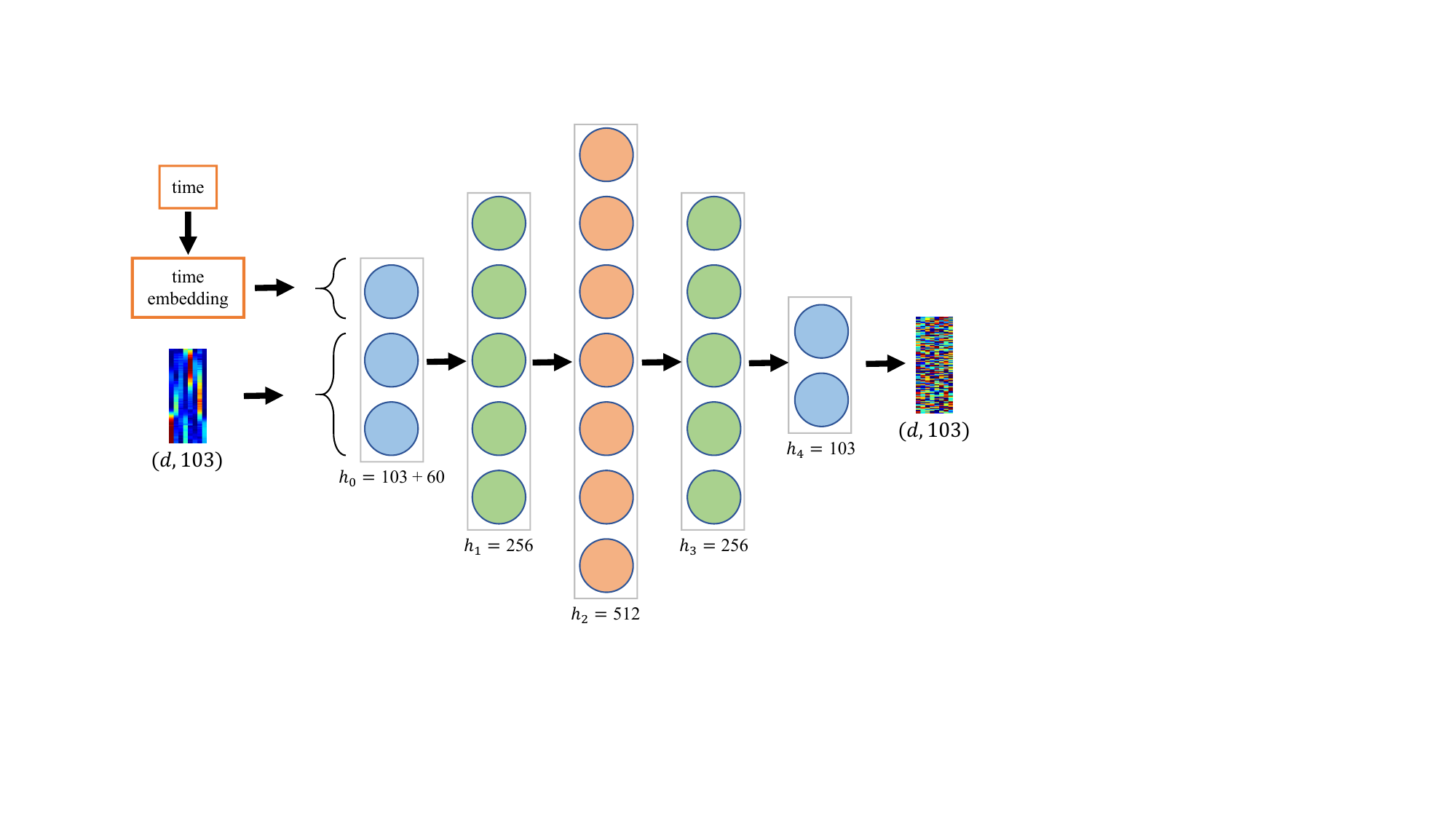}
    \vspace{-15mm}
    \caption{Spectral Network}
  \end{subfigure}
  
  \caption{The detailed network structures using the Pavia dataset as an example.}
  \label{Structures}
\end{figure}

The detailed network structure is illustrated in Figure ~\ref{Structures}. To explore the most cost-effective spatial-spectral network, we conducted ablation experiments on these two networks. Specifically, for the spatial network, we performed an ablation study on the number of channels within the network, controlled by the \texttt{channel\_mult} parameter. For the spectral network, we examined the effect of varying the dimensions of hidden layers, determined by the \texttt{hidden\_dims} parameter. The experimental results are summarized in Table~\ref{net}. Considering both accuracy and model size, we selected the spatial network with \texttt{channel\_mult} = \{1,2,3,4\} and the spectral network with \texttt{hidden\_dims} = \{256,512,256\}. (\textbf{Note:} Based on our existing experimental results, the ablation study of the spatial network is conducted with the optimal spectral network configuration by default, and vice versa for the spectral network ablation study.)

\begin{table}[htbp]
    \centering
    \caption{Ablation analysis of spectral and spatial networks.}
    \label{net}  
    \begin{scriptsize}
    \begin{tabular}{ccccc}
        \toprule
        Networks & channel\_mult/hidden\_dims & PSNR  & Params (M)\\
        \midrule
        \multirow{3}{*}{Spa. Net}
        &1,1,2,4  &41.55&15.60\\
        &1,2,3,4  &42.33&21.46\\
        &1,2,4,8  &42.40&57.03\\
        \midrule
        \multirow{3}{*}{Spe. Net}
        &256,256          &41.64&0.19\\
        &256,512,256      &42.33&0.39\\
        &256,512,512,256  &42.36&0.65\\
        \bottomrule
    \end{tabular}
    \end{scriptsize}
\end{table}

\section{Impact of Two Components}
\label{sec:tc}



Since our method simultaneously updates both the spectral and spatial components, we conduct an ablation study to investigate their individual contributions by evaluating the cases where only the spectral component or only the spatial component is updated. Specifically, when updating only the spectral component, we fix the spatial component by deriving \(\mathcal{A}\) from \(\mathcal{Y}\) (HR-MSI) using the Archetypal Analysis unmixing method \cite{cutler1994archetypal}. Conversely, when updating only the spatial component, we fix the spectral component by estimating \(\mathbf{E}\) from \(\mathcal{X}\) (LR-HSI) following the approach of PLRDiff \cite{rui2023unsupervised}. The final results are presented in Table~\ref{t}, showing that when only the spectral component is updated, the fusion accuracy significantly decreases. Additionally, when only the spatial component is updated, the PSNR of the fusion results remains approximately 2 dB lower than when both components are updated simultaneously.

\begin{table}[htbp]
    \centering
    \caption{Ablation study on two components.}
    \label{t} 
    \begin{scriptsize}
    \begin{tabular}{ccc cc}
        \toprule
        Update & PSNR\(\uparrow\) & SAM\(\downarrow\) & EGARS\(\downarrow\) & SSIM\(\uparrow\)\\
        \midrule
        Only \(\mathbf{E}\)        & 21.77  & 11.82&13.59& 0.296      \\
        Only \(\mathcal{A}\)       & 40.53  & 3.28 &1.77 & 0.969      \\
        \(\mathbf{E}\) \& \(\mathcal{A}\)        & 42.33  & 2.64 &1.49 & 0.977      \\
        \bottomrule
    \end{tabular}
    \end{scriptsize}
\end{table}

\vspace{-5pt}
\section{Sensitivity Analysis of Hyperparameters}
\label{sec:sk}


We present the parameter analysis of the balance weights \(\lambda_1\) and \(\lambda_2\) in Table~\ref{lambda}. Since the performance is optimal and relatively stable around \((1,1)\), we set both values to 1.

\begin{table}[h]
\centering
\caption{Sensitivity analysis of the parameters $\lambda_1$ and $\lambda_2$. }
\label{lambda}
\begin{scriptsize}
\begin{tabularx}{0.37\textwidth}{c|ccccc}
\specialrule{\heavyrulewidth}{0pt}{0pt}
\backslashbox{$\lambda_2$}{$\lambda_1$} & 0.1   & 0.5  & 1   & 2  & 5 \\
\hline
0.1    & 42.08 & 42.22 & 42.33 & 41.91 & 41.44  \\
0.5    & 42.15 & 42.31 & 42.25 & 41.90 & 41.63  \\
1      & 42.26 & 42.30 & 42.33 & 41.99 & 41.59  \\
2      & 42.04 & 42.06 & 42.13 & 42.33 & 41.40  \\
5      & 41.61 & 41.80 & 41.41 & 41.19 & 40.00  \\
\specialrule{\heavyrulewidth}{0pt}{0pt}
\end{tabularx}
\end{scriptsize}
\end{table}

\end{document}